\pdfoutput=1

\documentclass[11pt]{article}
\usepackage{latex/acl}

\usepackage{tablefootnote}
\usepackage{threeparttable}
\usepackage{longtable}
\definecolor{softgreen}{RGB}{34, 149, 34}

\definecolor{softred}{RGB}{232, 34, 34}
\newcommand{\cmark}{\textcolor{softgreen}{\ding{51}}} 
\newcommand{\xmark}{\textcolor{softred}{\ding{55}}} 
\usepackage{enumitem}
\usepackage{times}
\usepackage{pifont}
\usepackage{latexsym}
\usepackage{graphicx}
\usepackage{multirow}
\usepackage{tabularx}
\usepackage{amssymb}
\usepackage{booktabs}

\usepackage[normalem]{ulem}
\usepackage{colortbl}
\usepackage[most]{tcolorbox}

\definecolor{as}{RGB}{250, 125, 160}
\definecolor{for}{RGB}{160, 250, 78} 

\definecolor{as}{RGB}{241,156,153}
\definecolor{for}{RGB}{151,208,119} 
\definecolor{as}{RGB}{248, 206,204}
\definecolor{for}{RGB}{213,232,212} 
\definecolor{correctgreen}{rgb}{0.0, 0.5, 0.0} 
\definecolor{wrongred}{rgb}{0.8, 0.0, 0.0} 
\usepackage[T1]{fontenc}

\usepackage[utf8]{inputenc}

\usepackage{microtype}

\usepackage{inconsolata}
\usepackage{soul}

\usepackage{float}
\usepackage{xspace}
\usepackage{CJKutf8}
\tcbset{
  aibox/.style={
    width=474.18663pt,
    top=10pt,
    colback=white,
    colframe=black,
    colbacktitle=black,
    enhanced,
    center,
    attach boxed title to top left={yshift=-0.1in,xshift=0.15in},
    boxed title style={boxrule=0pt,colframe=white,},
  }
}
\newtcolorbox{AIbox}[2][]{aibox,title=#2,#1}

\begin{document}
\begin{CJK}{UTF8}{gbsn}

\title{LLMs {\color{red}\st{as}} { \color{blue} for} Doctors: Leveraging  Medical LLMs to {\color{blue} Assist Doctors},\\ Not {\color{red}Replace Them}}
\author{%
\textbf{Wenya Xie}, 
\textbf{Qingying Xiao},
\textbf{Yu Zheng},
\textbf{Xidong Wang}, 
\textbf{Junying Chen},
\textbf{Ke Ji},\\
\textbf{Anningzhe Gao}, 
\textbf{Xiang Wan}, 
\textbf{Feng Jiang$^*$},
\textbf{Benyou Wang}\thanks{%
Feng Jiang and Benyou Wang are the corresponding authors.}\\
Shenzhen Research Institute of Big Data\\
The Chinese University of Hong Kong, Shenzhen\\
\texttt{jeffreyjiang@cuhk.edu.cn} \texttt{wangbenyou@cuhk.edu.cn} \\
}
\maketitle

\begin{abstract}

The recent success of Large Language Models (LLMs) has had a significant impact on the healthcare field, providing patients with medical advice, diagnostic information, and more. However, due to a lack of professional medical knowledge, patients are easily misled by generated erroneous information from LLMs, which may result in serious medical problems. To address this issue, we focus on tuning the LLMs to be medical assistants who collaborate with more experienced doctors. We first conduct a two-stage survey by inspiration-feedback to gain a broad understanding of the real needs of doctors for medical assistants. Based on this, we construct a Chinese medical dataset called DoctorFLAN to support the entire workflow of doctors, which includes 92K Q\&A samples from 22 tasks and 27 specialists. Moreover, we evaluate LLMs in doctor-oriented scenarios by constructing the DoctorFLAN-\textit{test} containing 550 single-turn Q\&A and DotaBench containing 74 multi-turn conversations. The evaluation results indicate that being a medical assistant still poses challenges for existing open-source models, but DoctorFLAN can help them significantly. It demonstrates that the doctor-oriented dataset and benchmarks we construct can complement existing patient-oriented work and better promote medical LLMs research.
\end{abstract}

\section{Introduction}

Current large language models for healthcare applications\cite{yunxiang2023chatdoctor, singhal2023towards, wang2023clinicalgpt, chen2023huatuogpt},  primarily focus on providing online consultations to patients\cite{zhang2023huatuogpt, wang2023huatuo, chen2023bianque}. These models have demonstrated substantial potential and value, not only in reducing healthcare costs but also in enhancing the convenience of medical consultations.

However, several challenges hinder their deployment in real-world scenarios\cite{umapathi2023med, wang2023cmb}. Firstly, patients often lack the professional knowledge necessary to verify the accuracy of the model's outputs. This can pose significant risks when these models provide incorrect or misleading advice. 
Furthermore, issues such as model hallucination currently impede their practical application~\cite{fan2023quantifying}.

Recognizing these challenges, we advocate for repositioning LLMs in healthcare as supportive collaborators alongside experienced doctors who can identify and correct inaccuracies in the model’s outputs, ensuring safer implementation.
However, effectively integrating LLMs into medical practice necessitates the development of new tasks that go beyond the online consultation LLMs for patients. 
For example, doctors may need assistance with differential diagnosis, which is not required for patients in online consultation scenarios.

To achieve this goal, we conduct a two-stage survey using an inspiration-feedback method to gain a comprehensive understanding of the real needs of doctors for medical assistants. 
From the survey, we identify four tasks that doctors consider most effective under LLM's assistance, positioning them as ideal starting points in practice.

Building upon these insights, we develop DoctorFLAN, a comprehensive Chinese medical dataset that captures the full spectrum of the diagnostic process. 
Additionally, to assess the real-world clinical capabilities of these assistant LLM models, we designate DoctorFLAN-\textit{test} for single-turn evaluations and introduce DotaBench, a three-turn evaluation that can mimic actual medical interactions.

After that, we benchmark the existing medical LLMs and our model DotaGPT trained on DoctorFLAN automatically and manually. Our evaluation results indicate that existing online consultations medical models, struggle with the diverse and complex tasks of real-world assistant roles. 
Conversely, models trained on DoctorFLAN exhibit robust performance across tasks in DoctorFLAN-\textit{test} and DotaBench.
Moreover, based on our findings, some of the tasks prioritized by doctors (e.g., \textit{Preoperative Education}) in the previous survey are promising candidates for initial deployment due to the models' impressive performance in these areas.

The contributions of this work are manifold. 1) We first propose \colorbox{for}{\strut LLMs for Doctors}, a paradigm shift in the application of LLMs in healthcare, and conduct a two-stage survey to identify the real needs of clinical doctors in their daily practice.
2) We build the DoctorFLAN, containing about 92K samples covering 22 tasks in the entire workflow of doctors, to fill the gap for training LLMs' assistance to doctors with professional responses.
3) We develop the first expert-involved benchmark to evaluate LLMs in doctor-oriented scenarios with the single-turn DoctorFLAN-test and the multi-turn DotaBench, complementing the existing evaluation designed for patient-oriented scenarios.

\section{LLMs as Doctors or for Doctors?}
LLMs in healthcare applications fulfill two crucial roles: \colorbox{as}{\strut {LLMs as Doctors}} that autonomously provide medical consultants for patients and \colorbox{for}{\strut LLMs for Doctors}, which assist doctors in their daily work~\citep{knoll2022user, cai2022generation,krishna2020generating}. 
In this section, we first discuss the challenges with \colorbox{as}{\strut {LLMs as Doctors}} and advocate for a strategic shift toward safer and more effective applications.

\subsection{Challenges with LLMs as Doctors}
Popular Medical LLMs such as ChatDoctor~\cite{yunxiang2023chatdoctor} and HuatuoGPT~\cite{zhang2023huatuogpt} are designed to autonomously provide medical consultations. These models are trained on extensive patient-doctor dialogues to simulate medical advice. 
However, current implementations of such medical models exhibit relatively low accuracy rates~\citep{fan2023quantifying}. 
Exposing them to patients who don't have professional medical knowledge directly may pose significant risks as they could receive misleading responses.
For example, a patient consulting an LLM about severe stomachache and fever might receive incorrect advice to simply rest, potentially overlooking critical conditions like appendicitis that require immediate medical intervention.

\subsection{LLMs for Doctors}
Recognizing the discriminative ability of doctors with expertise against the errors caused by models, we propose \colorbox{for}{\strut LLMs for Doctors}, which have more practical significance and safer implementation by positioning the LLM as an assistant. 
Doctors can quickly identify and correct errors in diagnosis made by \colorbox{for}{\strut LLMs for Doctors}, thereby safeguarding patient health through critical intervention.

However, to develop LLMs that truly support doctors, we need to address significant gaps in existing datasets for model training, which we categorize as \colorbox{as}{\strut LLMs as Doctors}  datasets. As shown in Table \ref{tab:dataset_comparison}, current \colorbox{as}{\strut LLMs as Doctors} datasets often fail to capture the full scope of medical practice\footnote{See details in Appendix \ref{appendix:data_comparison}.} (e.g., OMCD), do not align well with real-world clinical scenarios (e.g., KG), or lack comprehensive responses (e.g., MCQA).

\begin{table}[ht]
\centering
\setlength{\tabcolsep}{3pt}
\begin{threeparttable}[b]
\resizebox{\linewidth}{!}{%
\begin{tabular}{lllccc}
\toprule
\multirow{2}{*}{\textbf{Paradigm}}  & \multirow{2}{*}{\textbf{Dataset}}  &    \textbf{Data}  &  \textbf{Entire} & \textbf{Clinical } & \textbf{Comprehensive } \\
  &   & \textbf{Type} &  \textbf{Workflow} & \textbf{ Scenarios} & \textbf{ Responses} \\
\midrule
& Huatuo-26M &  OMCD & \xmark  & \cmark& \cmark \\
& HealthCareMagic100k  & OMCD & \xmark   & \cmark& \cmark\\
& ChatDoctor10k &  OMCD & \xmark & \cmark& \cmark\\
& MedDialog  & OMCD & \xmark   & \cmark& \cmark\\
LLMs as& webMedQA& OMCD & \xmark & \cmark & \cmark\\
Doctors& KUAKE-QIC& OMCD & \xmark & \cmark & \cmark\\
& CMeKG & KG &  \cmark  & \xmark& \xmark\\
& cMedQA2 & MCQA & \cmark & \cmark& \xmark \\
& MedMCQA & MCQA & \cmark & \cmark & \xmark  \\
\hline
LLMs for & DoctorFLAN & \multirow{2}{*}{DAQA} & \multirow{2}{*}{\cmark} & \multirow{2}{*}{\cmark}  & \multirow{2}{*}{\cmark}\\
 Doctors &\&DotaBench&& &&\\\hline
\end{tabular}%
}
\caption{Comparison of existing medical training datasets. OMCD represents Online Medical Consultant Dialogue; KG represents Knowledge Graph; MCQA represents multiple-choice Question Answer; DAQA represents Doctor-Assistant Question Answer. Detailed URLs are collected in Appendix \ref{appendix:urls}.}

\label{tab:dataset_comparison}
\end{threeparttable}
\end{table}

Given the fundamental shortcomings in these datasets, the models trained on them inherently struggle to effectively handle the diversity of real-world medical environments and deliver comprehensive responses\footnote{Refer to the performance of HuatuoGPT in Table \ref{tab:auto eval-DoctorFLAN-DotaBench}. HuatuoGPT is a typical \colorbox{as}{\strut LLMs as Doctors} model.}.
To address these issues, it is crucial to develop targeted datasets and establish comprehensive benchmark evaluations for \colorbox{for}{\strut LLMs for Doctors}, specifically designed to reflect the everyday clinical scenarios of doctors. 

\begin{figure}
    \centering
    \includegraphics[width=\columnwidth]{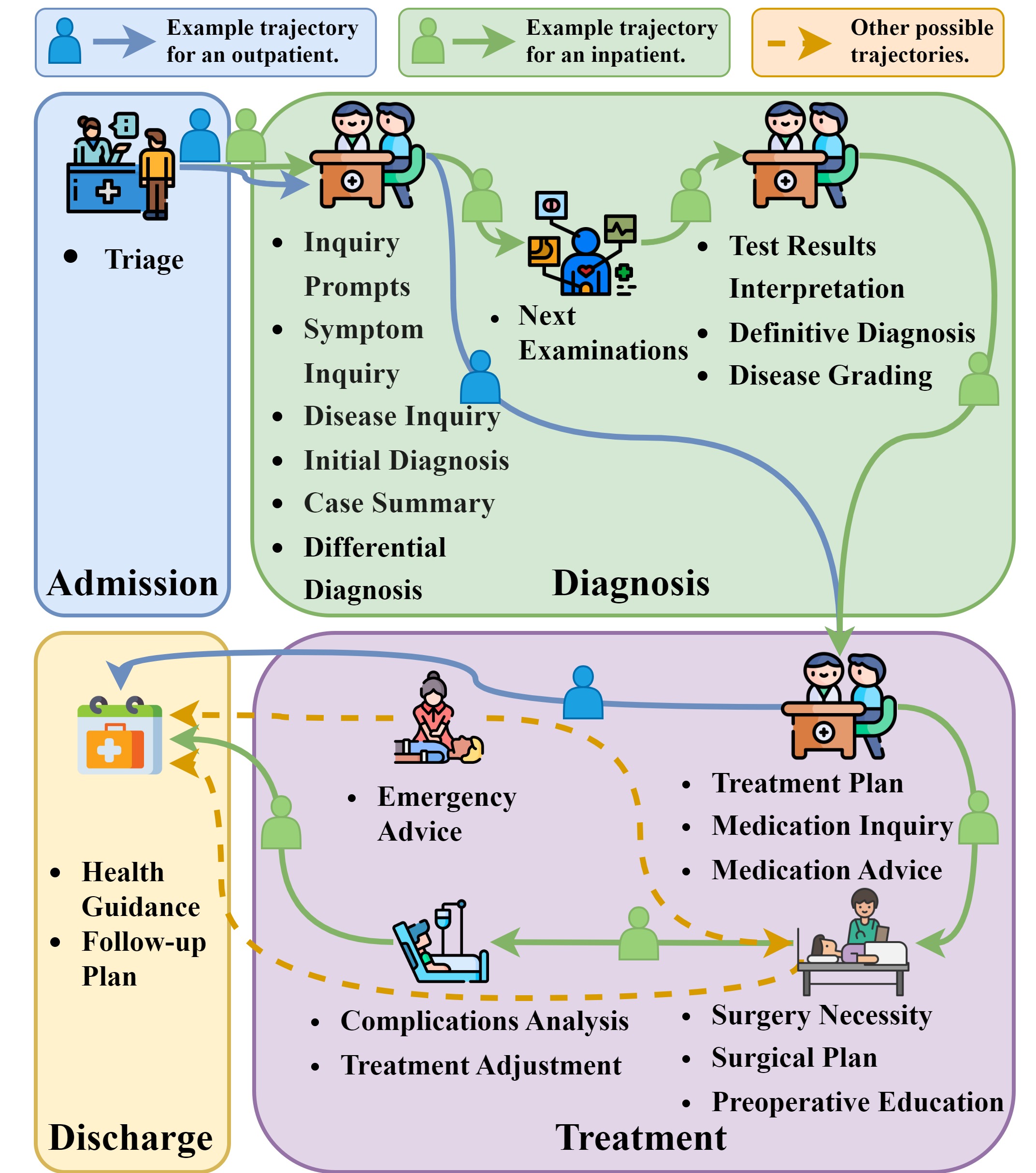}
    \caption{Overview of task categories in the LLMs for Doctors dataset across four phases: Admission, Diagnosis, Treatment, and Discharge, illustrating typical and alternative patient care pathways in solid and dashed lines, respectively.}
    \label{fig:full_process}
\end{figure}

\label{sec:survey}
\subsection{Identifying Doctors' Needs for LLMs}
To accurately capture the medical scenarios where LLMs can be most beneficial serving as \colorbox{for}{\strut LLMs for Doctors}, we design a two-stage heuristic-feedback task collection method to ensure the tasks involved in the medical workflow are professional and practical. Additionally, we conduct further analysis to determine which tasks are considered most effective for doctors when assisted by LLMs, aiming to prioritize these tasks for early model deployment.

\subsubsection{Task Collection}
Our initial step is to collaborate with experienced doctors to identify essential tasks in their daily workflows using a heuristic top-down method. Subsequently, we validate the universality of these tasks through an extensive user survey across multiple medical institutions with feedback.

\textbf{Heuristic Task Definition.} 
We organize a symposium with 16 healthcare professionals\footnote{Details of the healthcare professionals are provided in Appendix \ref{appendix:details_of_symposium}.} from four distinct specialties to establish task definitions based on heuristics, ensuring the tasks we identify are closely aligned with the practical needs of doctors. During the discussion, experts systematically categorize the medical workflow into 4 stages (pre-diagnosis, diagnosis, treatment, and post-treatment) with 22 tasks. It comprehensively covers all scenarios for the doctor's workflow, from the triage to the creation of follow-up plans, as shown in Figure \ref{fig:full_process}. Unlike typical online consultations, which primarily address pre-treatment issues, such as initial diagnoses, and disease prevention, our approach extensively covers a wide range of tasks that doctors might encounter in a hospital setting, encompassing both inpatient and outpatient tasks.

\begin{figure}[t]
\includegraphics[width=\columnwidth]{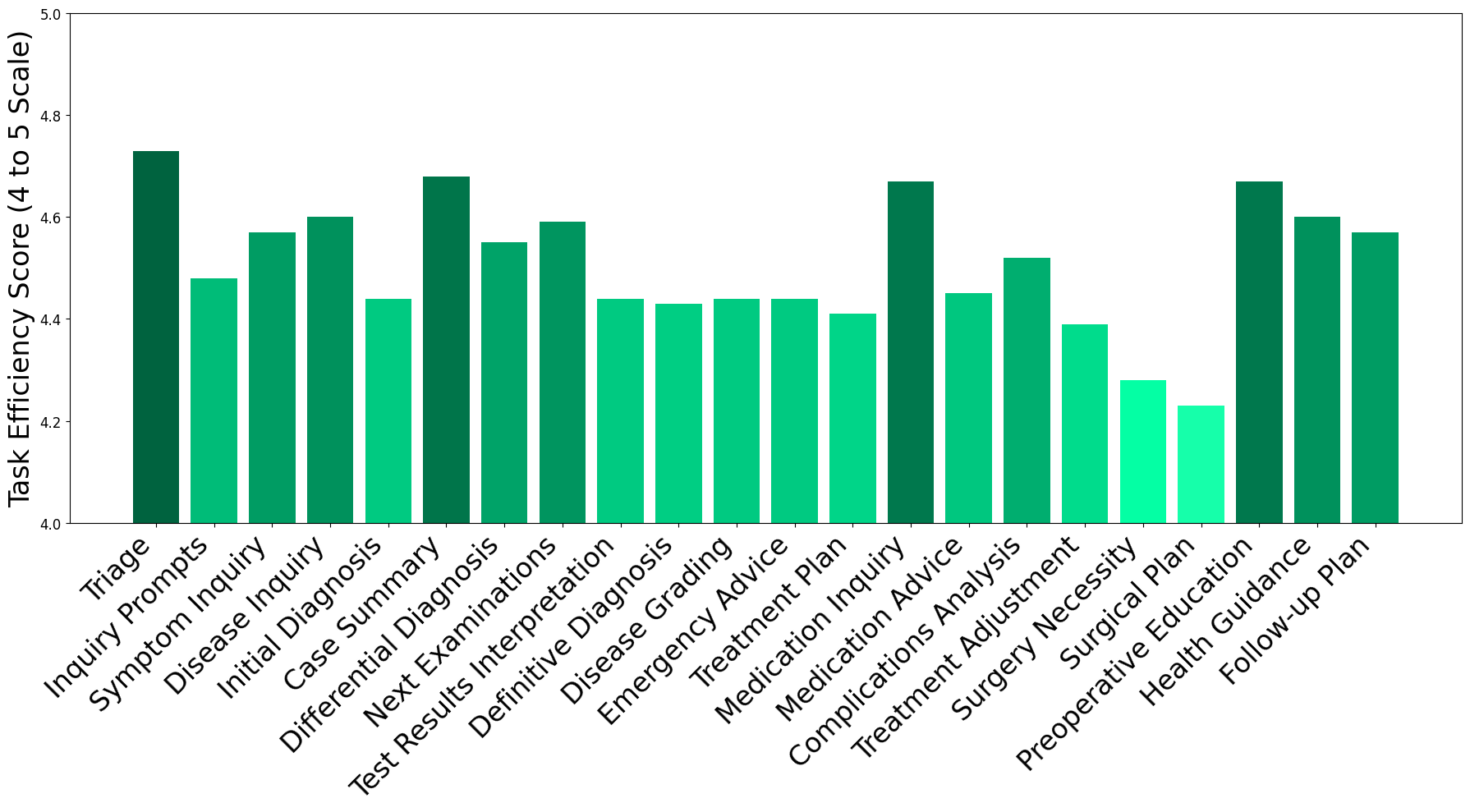}
\caption{Comparative assessment of task efficiency scores for each task according to our survey. The Task Efficiency Score quantifies the potential of various tasks to enhance operational efficiency in medical practice, reflecting improvements in time management, resource use, and overall workflow efficacy. }
    \label{fig: importance score}
\end{figure}

\textbf{Task Confirmation with Feedback.}
Then, we survey doctors from over a dozen top-tier hospitals with their feedback to validate the universality of these tasks. In the survey\footnote{For detailed methodology and the questionnaire used in the survey, see Appendix \ref{appendix:survey-questionnaire}.}, doctors are asked to rate the helpfulness of each task with LLMs' assistance, where a rating scale from 1 to 5 is given, with scores of 4 and 5 indicating high and critical applicability, respectively. In over 86\% (71/82) of the valid questionnaires collected, the higher scores achieved in all tasks demonstrate their effectiveness and importance as recognized by the surveyed doctors, as shown in Figure \ref{fig: importance score}.

\subsubsection{Analysis of Tasks Prioritized for Doctors} To identify which tasks should be prioritized for initial deployment of \colorbox{for}{\strut LLMs for Doctors}, we analyze the efficiency scores that doctors assign to each task in the survey.
It is noticed that tasks such as \textit{Triage}, \textit{Case Summary}, \textit{Medication Inquiry}, and \textit{Preoperative Education} have particularly high scores, which doctors consider highly beneficial due to the tasks' repetitive nature (e.g., case summarization, preoperative education), relatively low medical risk (e.g., triage), and extensive information requirements (e.g., medication inquiry).

\section{Dataset Construction}
To address these needs, we develop the DoctorFLAN and DotaBench for single-turn Q\&A and multi-turn conversation scenarios separately.

\subsection{DoctorFLAN}
To develop a dataset that comprehensively covers the entire medical workflow, we construct DoctorFLAN based on the tasks previously defined.
We collect source data from a wide range of medical data and complete the construction of the dataset through two steps: instruction normalization and response polishing. Additionally, we also ensure the high quality of the data by manually verifying some of the sampled data.

\textbf{Data Collection.} 
Our dataset primarily comprises multiple-choice questions sourced from Medtiku\footnote{https://www.medtiku.com/}, a repository of expert-level medical examination questions, and various online resources. These questions are selected due to their extensive coverage of professional medical knowledge and their relevance to real-world clinical scenarios. In addition, we supplement our dataset with encyclopedia entries\footnote{Hosted by 120 Ask, https://m.120ask.com/} for constructing inquiry tasks such as \textit{Medicine Inquiry}. We also incorporat a subset of the PromptCBLUE dataset~\citep{zhu2023promptcblue}, selected for its diverse tasks and high quality. 
After collecting all the original data, we perform deduplication and categorization using crafted regular expressions as part of the preprocessing stage.

\textbf{Refinement of Original Data.}
Given the lack of practicality of the original data, we adopt a refinement process consisting of two key components: refinement of instructions and enhancement of responses.
First, We ask doctors to manually draft the instructions for the 22 task categories to ensure that these instructions can reflect actual use scenarios. 
Second, in response to the newly refined instructions, we employ GPT-4~\footnote{GPT-4 is chosen for its superior performance among various models we tested.} to generate updated answers. 
Specifically, we provide GPT-4 with reference materials, such as the original multiple-choice question answers, to serve as a basis for generating responses. It enables GPT-4 to produce responses that are not only accurate but also tailored to the required context and depth, aligning with the practical demands and professional standards expected in the medical field.

\textbf{Data  Verification.} 
To ensure that the responses generated by the GPT-4 model are factually accurate and realistic, we use a structured review process in which a sample of 1050 responses (50 data points per item across 22 items) are reviewed by three medical professionals, each reviewing 350 items. 

The verification criteria include \textit{Correctness}, where a response is considered correct if it contains no factual errors, and \textit{Practicality}, where a response is deemed practical if it is more effective than the original answer. Our results demonstrate a high level of correctness (100\%) and practicality (99.9\%), underscoring the robustness of the DoctorFLAN. Further details (e.g., the graphical user interface used for verification and the detailed analysis of failure cases) are available in Appendix \ref{appendix:DoctorFLAN_checking}.

\subsection{DotaBench}

Extending the single-turn dataset DoctorFLAN, we introduce DotaBench to evaluate multi-turn dialogues among medical assistants. This enhancement reflects real-world needs where medical assistants must handle multiple rounds of Q\&A to thoroughly assess a condition, requiring continuous and consistent responses across interactions.

\textbf{Data Collection.} To ensure its high professionalism, we select CMB-Clin~\citep{wang2023cmb}, a multi-round question-answering dataset derived from real, complex medical records, as the data source for DotaBench.

\textbf{Refinement of Original Data.}
However, the samples from CMB-Clin, sourced from medical exams, exhibit a formal written style and often lack continuity between rounds of questioning.

To better simulate the multi-turn interactions typical between medical assistants and doctors, we engage medical professionals to rephrase these questions in a manner consistent with their routine inquiry processes. In handling multiple rounds of questioning, they are instructed to maintain logical sequence and coherence across three question items, thereby creating a more realistic simulation of conversational turns.

We employ the original Q\&A sets as a reference for evaluation. Unlike DoctorFLAN, which directly involves LLMs in data generation, DotaBench is crafted without LLM intervention, thereby eliminating the need for subsequent data verification and ensuring more controlled evaluation conditions.

\begin{table}
\centering
\resizebox{0.8\linewidth}{!}{
\begin{tabular}{@{}lccc@{}}
\toprule
    & \multicolumn{2}{c}{\textbf{DoctorFLAN}} & \textbf{DotaBench} \\
\midrule
\textbf{Type}          & \multicolumn{2}{c}{Single-turn}      & 3-turns           \\
\textbf{Split}         & train              & test               & test              \\
\textbf{Specialist}    & 27 &27              &  -          \\
\textbf{Task}          & 22    &22         &   -       \\
\textbf{\#Q/task}  & -*                  & 25                 & -                 \\
\textbf{\#Q in total}  & 91,330             & 550                & 74                \\
\bottomrule
\end{tabular}
}
\caption{The Statistics of DoctorFLAN and DotaBench Dataset. * Each task has a different number of questions, as detailed in Appendix \ref{appendix:data statistics}.}
\label{tab:dataset-summary}
\end{table}

\subsection{Data Statistic}
The statistical analysis of the DoctorFLAN and DotaBench datasets is presented in Tabel \ref{tab:dataset-summary}.
The DoctorFLAN dataset comprises 91,880 instances across 22 distinct tasks, involving 27 medical specialties in total as detailed in Appendix \ref{appendix:data statistics}, demonstrating the comprehensive coverage of DoctorFLAN in real clinical scenarios. 
In addition, we have extracted a subset of 25 instances from each task, referred to as DoctorFLAN-\textit{test} for evaluation. The DotaBench dataset includes 74 instances of 3-turn conversations.

\section{Experiments}
\subsection{Experimental Setup}
\subsubsection{Evaluation Models}
\textbf{Existing Baselines.}
To ensure a comprehensive understanding of how different models perform in real-world medical assistance, We evaluate the following Chinese medical LLMs on DoctorFLAN-\textit{test} and DotaBench: BianQue-2~\citep{chen2023bianque}, DISC-MedLLM~\citep{bao2023disc}, HuatuoGPT~\citep{zhang2023huatuogpt}, HuatuoGPT-II~\citep{chen2023huatuogpt}. In addition to these domain-specific models, we evaluate some models that perform strongly on generic tasks: Qwen-1.8B-Chat~\cite{bai2023qwen}, Baichuan-13B-Chat, Baichuan2-7B-Chat and Baichuan2-13B-Chat~\citep{yang2023baichuan}, Yi-6B-Chat and Yi-34B-Chat, Claude-3, GPT-3.5, GPT-4. More details about the selected models can be found in Appendix~\ref{appendix:details of existing models}. 

\textbf{Training Model.}
We also train models named DotaGPT on the DoctorFLAN training set as a baseline. We use DoctorFLAN training set to train  DotaGPT series on two backbones: Yi-6B and Baichuan2-7B-Base, respectively, as detailed in Appendix \ref{appendix:model_training}.

\subsubsection{Evaluation Method}
Considering both accuracy, reliability, and cost, our evaluation methodology incorporates both automatic and human evaluations.\footnote{For model evaluation, the decoding hyperparameters used across all models are max\_new\_tokens = 1024, top\_p = 0.7, temperature = 0.5, and repetition\_penalty = 1.1.}

\textbf{Automatic Evaluation.}
We use GPT-4\footnote{gpt-4-0125-preview} for evaluation, which has proven effective in the preliminary stages of development~\citep{liu2023gpteval}, rather than traditional metrics like F1 and BLEU, as these metrics often misjudge similar sentences that were semantically similar.
During the automatic evaluation phase, we adopt the evaluation methodology utilized by MT-Bench~\citep{zheng2024judging}, as the criteria and prompts designed for this scoring are detailed in Appendix \ref{appendix: Evaluation Prompt}. To ensure the accuracy of the GPT-4 scoring process, \textit{we supply the model with relevant contextual references} and direct it to base its scoring on these references. Furthermore, GPT-4 is instructed to utilize a set of well-defined criteria for evaluation, thereby enhancing the meaningfulness of the scores assigned.

\textbf{Human Evaluation.}
To validate the accuracy of the automatic evaluation, we conduct a human evaluation involving six medical experts assessing six models using DoctorFLAN-\textit{test}, and three medical experts evaluating three models on DotaBench. Details of the evaluation procedures and guidelines are provided in Appendix \ref{appendix: Human Eval Details}.

\begin{table*}[t]
\centering
\small
\resizebox{\textwidth}{!}{%
\begin{tabular}{lcccccc|c}
\toprule
\multirow{2}{*}{\textbf{Model}} & \multirow{2}{*}{\textbf{Size}} & \multicolumn{5}{c}{\textbf{DoctorFLAN-\textit{test}}} & \textbf{DotaBench} \\
 & & \textbf{Pre-Diagnosis} &  \textbf{During Diagnosis} & \textbf{Treatment} &  \textbf{Post-Treatment} &  \textbf{Average} & \textbf{Average} \\
\hline
\rowcolor{gray!10} \multicolumn{8}{c}{General Models} \\ \hline
Qwen-1.8B-Chat & 1.8B & 5.28 & 4.56 & 3.96 & 5.44 & 4.48 & 5.77\\
Baichuan-13B-Chat & 13B & 6.20 & 6.51 & 6.31 & 7.55 & 6.57 & 8.17 \\
Baichuan2-7B-Chat & 7B & 6.32 & 6.36 & 6.34 & 7.70 & 6.59 & 8.33 \\
Baichuan2-13B-Chat & 13B & 6.76 & 6.85 & 6.94 & 7.81 & 7.04 & 8.33 \\
Yi-6B-Chat & 6B & 7.00 & 6.83 & 6.83 & 7.66 & 6.98 & 8.27\\
Yi-34B-Chat& 34B  & 7.36 & 7.38 & 7.95 & 8.78 & 7.80 & 8.81\\
\hline
\rowcolor{gray!10} \multicolumn{8}{c}{Medical Models} \\ \hline
BianQue-2 & 6B & 5.56 & 3.27 & 3.65 & 4.78 & 3.72 & 4.92 \\
DISC-MedLLM& 13B & 5.56 & 4.23 & 3.54 & 5.14 & 4.24 & 5.81\\
HuatuoGPT & 7B  & 5.32 & 4.24 & 3.72 & 4.92 & 4.29 & 6.78\\
HuatuoGPT-II& 7B  & 7.60 & 7.02 & 6.69 & 7.42 & 7.03 & 8.55\\
\rowcolor{gray!10} DotaGPT (Yi-6B)& 6B & 8.32 & 7.62 & 7.68 & 8.44 & 7.81\textsubscript{{\color{blue} $\uparrow$ 11.9\%}} & 9.05 \textsubscript{{\color{blue} $\uparrow$ 9.4\%}}\\
\rowcolor{gray!10} DotaGPT (Baichuan2-7B-Base)& 7B & 8.48 & 8.01 & 8.23 & 8.80 & 8.25\textsubscript{{\color{blue} $\uparrow$ 25.2\%}} &9.00 \textsubscript{{\color{blue} $\uparrow$ 8.0\%}}\\
\hline
GPT-3.5 & N/A & 6.40 & 6.85 & 6.26 & 6.74 & 6.64 & 8.64\\
Claude-3 & N/A & 7.80 & 8.38 & 8.28 & 8.76 & 8.38 & 9.46\\
GPT-4 & N/A & 8.00 & 8.41 & 8.28 & 9.04 & 8.42 & 9.41\\ \hline
Average Score & - & 6.90 & 6.59 & 6.50 & 7.43 & - & - \\\hline
\end{tabular}
}
\caption{Automatic Evaluation Results on DoctorFLAN-\textit{test} and DotaBench. For detailed task-by-task results on DoctorFLAN-\textit{test}, see Appendix \ref{appendix:automatic_doctorflan_all}; for results by turn on DotaBench, see Appendix \ref{appendix:details_of_dotabench}.}
\label{tab:auto eval-DoctorFLAN-DotaBench}
\end{table*}
\subsection{Automatic Evaluation Results}

\begin{table}[t]
\centering
\small
\resizebox{0.9\linewidth}{!}{
\begin{tabular}{lc}
\toprule
\textbf{Models} & \textbf{Average Score} \\
\midrule
BianQue-2 & 4.58 \\
HuatuoGPT & 4.97 \\
DISC-MedLLM & 5.36 \\
Baichuan2-7B-Chat & 6.69 \\
DotaGPT (Baichuan2-7B-Base) & 7.83 \\
GPT-4 & 8.06 \\
\bottomrule
\end{tabular}
}
\caption{Human Evaluation Results on DoctorFLAN-\textit{test}. For detailed task-by-task results, see Appendix \ref{appendix: Human Eval Details}.}
\label{tab:average_scores}
\end{table}

\subsubsection{On the DoctorFLAN-\textit{test}}
The performance of general models and medical models are shown in Table \ref{tab:auto eval-DoctorFLAN-DotaBench}. It is also noteworthy that BianQue-2 and HuatuoGPT, typical models representing \colorbox{as}{\strut LLMs as Doctors}, exhibit poor performance. Analysis by phase reveals that models generally score lower during the diagnosis and treatment phases tasks compared to an average score of 6.9 in the pre-diagnosis phase tasks and 7.43 in the post-treatment phase tasks for all models. 
Interestingly, \textit{medical models do not always exhibit stronger performance than general LLMs}. For instance, DISC-MedLLM(4.24) underperforms compared to its general counterpart Baichuan-13B-Chat(6.57), underscoring the risks of excessive specialization, indicating that a balance between specialization and general adaptability is essential for broader applicability.

In contrast, our DotaGPT variants, fine-tuned on the DoctorFLan dataset, demonstrate significant performance improvements over their respective chat model counterparts. Specifically, the variant fine-tuned with Baichuan-2-7B-base shows a substantial improvement of 25.2\%, reaching a performance score of 8.25. Similarly, the DotaGPT variant fine-tuned with Yi-6B outperforms the Yi-6B-Chat by 11.9\%.

The improvement on both backbones highlights the effectiveness of DoctorFLAN and brings our models' performance close to those of leading proprietary models such as Claude-3 and GPT-4.

\begin{table}[t]
\centering
\small
\begin{tabular}{lc}
\toprule
\textbf{Model}  & \textbf{Average Score}  \\ \midrule
HuatuoGPT-II      & 7.98 \\
Baichuan2-7B-Chat & 8.25 \\
DotaGPT (Baichuan2-7B-Base)  & 8.54\\ 
\bottomrule
\end{tabular}%
\caption{Human Evaluation Results on the DotaBench.}
\label{tab: expert eval}
\end{table}

\subsubsection{On DotaBench}
As shown in Table \ref{tab:auto eval-DoctorFLAN-DotaBench}, our DotaGPT variants significantly outperform models of comparable size on DotaBench. Surprisingly, it even outperforms the Yi-34B-Chat, which is significantly larger in model size. 
This strong performance of the DotaGPT variants on DotaBench, which includes out-of-domain (OOD) challenges, underscores their robust training on the DoctorFLAN-\textit{training} dataset. This training not only boosts their capabilities within specific domain tasks but also effectively generalizes to out-of-domain contexts. 

\begin{figure}
    \centering
    \includegraphics[width=0.8\columnwidth]{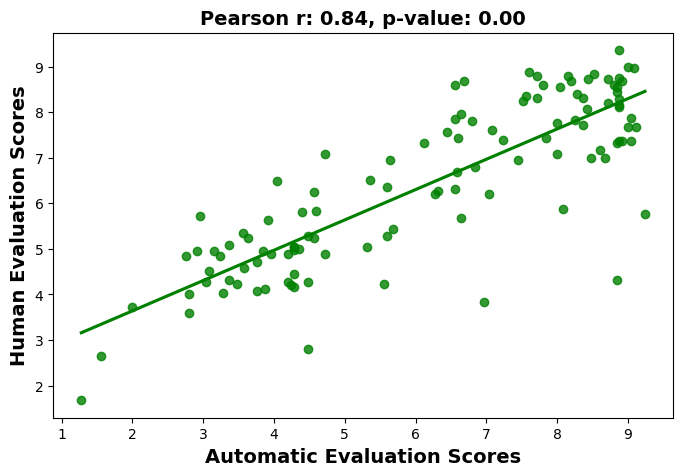}
    \caption{Pearson correlation between human and automatic evaluations on DoctorFLAN-\textit{test}, illustrating task-level consistency.}
    \label{fig:correlation}
\end{figure}

\begin{table}[ht]
\centering
\setlength{\tabcolsep}{3pt}
\resizebox{\linewidth}{!}{%
\begin{tabular}{lccccc}
\hline
\multirow{2}{*}{\textbf{Model}} & \multirow{2}{*}{\textbf{Triage}} & \textbf{Case} & \textbf{Medication} & \textbf{Preoperative} & \multirow{2}{*}{\textbf{Average}} \\
&& \textbf{Summary} & \textbf{Inquiry} & \textbf{Education} & \\\hline
\rowcolor{gray!10} \multicolumn{6}{c}{General Models} \\
\hline
Qwen-1.8B-Chat & 5.28 & 5.64 & 3.56 & 5.00 & 4.87 \\
Baichuan-13B-Chat & 6.20 & 8.52 & 6.67 & 7.67 & 7.27 \\
Baichuan2-7B-Chat & 6.32 & 8.84 & 5.60 & 7.44 & 7.05 \\
Baichuan2-13B-Chat & 6.76 & 8.93 & 6.56 & 8.49 & 7.69 \\
Yi-6B-Chat & 7.00 & 8.28 & 6.96 & 8.28 & 7.63 \\ 
Yi-34B-Chat & 7.36 & 8.56 & 5.88 & 9.28 & 7.77 \\
\hline
\rowcolor{gray!10} \multicolumn{6}{c}{Medical Models} \\
\hline
BianQue-2 & 5.56 & 1.28 & 3.08 & 1.56 & 2.87 \\
DISC-MedLLM & 5.56 & 3.20 & 3.96 & 4.24 & 4.24 \\
HuatuoGPT & 5.32 & 4.48& 4.60 & 4.56& 4.74\\
HuatuoGPT-II & 7.60 & 7.04 & 6.88 & 7.56 & 7.27 \\
\rowcolor{gray!10} DotaGPT (Yi-6B) & 8.32 & 6.48 & 7.48 & 8.84 & 7.78 \\
\rowcolor{gray!10} DotaGPT (Baichuan2-7B-base) & 8.48 & 6.96 & 8.00& 9.04 & 8.12 \\ 
\hline
GPT-3.5 & 6.40 & 9.08 & 6.32 & 7.60 & 7.35 \\
Claude-3 & 7.80 & 9.32 & 7.36 & 9.00 & 8.37 \\
GPT-4 & 8.00 & 9.24 & 6.60 & 8.88 & 8.18 \\
\bottomrule
\end{tabular}
}
\caption{Performance of General and Medical Models on Tasks Prioritized for
Doctors within the DoctorFLAN-\textit{test}.}
\label{tab:doctor_focus}
\end{table}

\subsection{Human Evaluation Results}

\textbf{Overall Performance.}
As indicated in Table \ref{tab:average_scores}, DotaGPT (Baichuan2-7B-Base) demonstrates exceptional performance in manual evaluations, substantially surpassing both HuatuoGPT and Baichuan2-7B-Chat and closing to GPT-4 on the DoctorFLAN-\textit{test}. Furthermore, on the DotaBench, DotaGPT (Baichuan2-7B-Base) not only outperforms its general counterpart, Baichuan2-7B-Chat but also exceeds another SOTA model, HuatuoGPT-II, achieving the highest scores among the three models evaluated.

\textbf{Correlation with Automatic Evaluations.}
To verify the reliability of our evaluation methods, we conducted a task-level correlation analysis between human and automatic evaluations, aligned with the structured review process by healthcare professionals who evaluated specific tasks across models. Our analysis, covering 132 data points, reveals a Pearson correlation coefficient of 0.84, indicating strong consistency between evaluation modes, as shown in Figure \ref{fig:correlation}.

\section{Analysis}
To expedite the deployment of LLM assistants in clinical settings, we first analyze four key tasks identified by doctors in the previous survey (see Section \ref{sec:survey}). We then conduct a detailed case study to further illustrate the advantages of models trained using our DoctorFLAN dataset. Additionally, we compare our evaluation results with other benchmarks to demonstrate the necessity of our specialized assessments.
\subsection{Performance on Tasks Prioritized for Doctors}
We analyze model performance on tasks specifically relevant to doctors, as detailed in Table \ref{tab:doctor_focus}. 
In the \textit{Case Summary} task, models such as Yi-34B-Chat and DotaGPT (Baichuan2-7B-Base) excel due to advanced training datasets and modeling techniques, showcasing enhanced text generation capabilities. They demonstrate high accuracy and are suitable for immediate use in clinical settings to improve workflow efficiency. Conversely, the \textit{Medication Inquiry} task presents significant challenges to several models, requiring precise medical knowledge. While specialized medical models like HuatuoGPT-II and DotaGPT (Baichuan2-7B-Base) outperform general counterparts, the need for domain-specific accuracy suggests that these models should only be deployed once they have matured further, or when used, their outputs should be verified with external tools to ensure reliability. This cautious approach is necessary to prevent errors that could impact patient safety.

\begin{table}[ht]
\centering
\setlength{\tabcolsep}{3pt}
\resizebox{\linewidth}{!}{
\begin{tabular}{lccc}
\toprule
\textbf{Model}  & \textbf{CMB-Exam} & \textbf{DoctorFLAN-\textit{test}} & \textbf{Rank Change}\\
\hline
\rowcolor{as} Yi-34B-Chat     & 69.17 & 7.80 (3) & -2 \\
\rowcolor{as}Yi-6B-Chat      & 65.87 & 6.98 (5) & -3 \\
\rowcolor{for}GPT-4           & 59.46 & 8.42 (1) & +2\\
HuatuoGPT-II    & 59.00 & 7.03 (4) & +0 \\
\rowcolor{for}DotaGPT (Baichuan2-7B-Base) & 52.42 & 8.25 (2) &+3\\
Baichuan2-7B-Chat & 43.33 & 6.59 (6)& +0 \\
GPT-3.5         & 38.31 & 6.64 (7) &+0\\
\rowcolor{for}DISC-MedLLM     & 32.47 & 4.24 (10) & +2\\
HuatuoGPT       & 30.73 & 4.29 (9) &+0 \\
\rowcolor{as}Qwen-1.8B-Chat  & 27.43 & 4.48 (8) & -2\\
BianQue-2       & 3.92 & 3.72 (11) & +0 \\
\bottomrule
\end{tabular}
}
\caption{Comparative performance of medical LLMs on the CMB-Exam and DoctorFLAN-\textit{test}(rankings). CMB-Exam scores are cited from \citep{wang2023cmb}.}
\label{tab:MCQA_results}
\end{table}

\begin{table*}[ht]
\centering
\resizebox{\linewidth}{!}{
{\scriptsize 
\begin{tabular}{l}
\hline
\textbf{Query:} \\
Based on the patient's history, symptoms, and examination results, list several possible diagnoses first, \\
then explain why these diagnoses may correspond to the patient's condition and how to confirm the final diagnosis through further examination or observation. \\
<Case description>: \\
The patient was a male, 11 months old, whose parents complained of a 2-week-old \textbf{white reflection in the centre of the right eye, with no redness of the eye and no}\\ \textbf{strabismus}. There was no family history of this condition, and the patient denied any history of trauma. He had a full-term delivery with no history of oxygen \\intake and no history of birth injury.\\\hline
\textbf{Golden:}\\
Based on the history and symptoms of the child, the following are a few possible initial diagnoses:\\1. Congenital cataract - The child is young and has no other medical history such as prematurity...\\2. Retinoblastoma (Retinoblastoma) - Although there is no such case in the family...\\
...\\
Why these diagnoses may be consistent with the patient's condition: \\- Congenital cataract: fundus reflections may be produced by a light source hitting the cloudy lens.. \\- Retinoblastoma: white fundus reflections (also known as cat's eye reflex) in children are typical of retinoblastoma. \\- Consider developmental anomalies such as PHPV: usually occurring unilaterally...\\\hline
\textbf{DotaGPT (Baichuan2-7B-Base):}\\
Based on the information you have provided in your medical history, here are some possible preliminary diagnoses and their plausibility:\\
\#\#\#\# Possible Diagnoses:\\
1. \color{correctgreen}{Congenital Cataract}\cmark:\\ - Rationale: \color{correctgreen}{A cataract usually presents as a white reflection in the pupil area or at the location of the lens...} \cmark  \\ - FURTHER EXAMINATION: Ultrasound may be performed to assess whether the structure and position of the lens is normal...\\ 
2. \color{correctgreen}{Retinoblastoma (Retinoblastoma)}\cmark:\\ - Rationale: In infants and children, retinoblastoma is the most common malignant tumour of the eye. \color{correctgreen}{Although usually associated with symptoms such as redness} \\\color{correctgreen}{of the eye and strabismus, early lesions may also present only as white reflections.}\cmark\\...\\\hline
\textbf{HuatuoGPT:}\\
Based on the history you have provided, here are some possible diagnoses:\\
1. Conjunctivitis: Conjunctivitis is a common eye condition usually caused by a viral or bacterial infection... \\
2. Cataracts: Cataracts are a common eye disease in the elderly, usually resulting in clouding of the lens and affecting vision. \\Cataracts are \color{wrongred}{also a possible diagnosis if your child has no other symptoms such as fever or cough}, such as headaches or vomiting.\xmark\\ 
3. Keratitis: Keratitis is an inflammatory disease ... Keratitis is \color{wrongred}{also a possible diagnosis if your child has no other symptoms such as fever or cough...}\xmark \\ 
4. Glaucoma: Glaucoma is a common eye disease ... Glaucoma is \color{wrongred}{also a possible diagnosis if your child has no other symptoms such as fever or cough...} \xmark\\ 
We can perform fundus examination, visual field examination and corneal staining to help determine the cause and confirm the diagnosis. \\\hline
\end{tabular}
}
}
\caption{The Case Study of Comparative Evaluation of Model Responses to a Medical \textit{Differential Diagnosis} Task. The \textcolor{correctgreen}{green} font indicates correct information, the \textcolor{wrongred}{red} font indicates incorrect information.}
\label{tab:case_study}
\end{table*}

\subsection{Case Study} 
As detailed in Table \ref{tab:case_study}, we examine a \textit{Differential Diagnosis} scenario to compare HuatuoGPT, representing \colorbox{as}{\strut LLMs as Doctors}, with DotaGPT, exemplifying \colorbox{for}{\strut LLMs for Doctors}. Despite HuatuoGPT's fluent responses, it struggled with critical patient details such as the \textit{
white reflection in the centre of the right eye}, showing a significant lack of contextual understanding necessary for accurately completing tasks required by doctors. Conversely, DotaGPT not only accurately diagnosed the condition but also provided detailed, professional reasoning, demonstrating its superior ability to integrate specific patient information and clinical context effectively. 

\subsection{Comparison with Other Benchmarks}
We compare the performance of models on our DoctorFLAN-\textit{test} with their results on the CMB-Exam \cite{wang2023cmb}, a medical multiple-choice benchmark, as detailed in Table \ref{tab:MCQA_results}. 
The comparison reveals inconsistent rankings between the two benchmarks, highlighting that success in multiple-choice scenarios does not necessarily predict similar outcomes in tasks that demand detailed narrative responses and complex reasoning typical of practical medical settings.

\section{Related Work}

\textbf{Medical LLMs for Application}
Current medical LLMs typically fine-tune general models using doctor-patient dialogues to enhance medical capabilities. Efforts like Bentsao{~\citep{wang2023huatuo}}, PMC-LLaMA{~\citep{wu2023pmc}}, Med-PaLM2{~\citep{singhal2023large}}, DISC-MedLLM{~\citep{bao2023disc}}, and HuatuoGPT{~\citep{zhang2023huatuogpt}} aim to deepen medical knowledge integration. However, while these models are primarily trained to simulate AI doctors, the development of models as doctor’s assistants remains underexplored.

\textbf{Clinical Benchmarking}
In medical LLMs, benchmarks like PubMedQA{~\citep{jin2019pubmedqa}}, MedQA{~\citep{jin2021disease}}, MultiMedQA{~\cite{singhal2023large}}, MedMCQA{~\citep{pal2022medmcqa}}, cMedQA2~\citep{zhang2018multi}, and CMB{~\citep{wang2023cmb}} have focused on assessing knowledge accuracy through multiple-choice questions. Additionally, evaluations like PromptCBLUE{~\citep{zhu2023promptcblue}} focus on assessing the abilities of LLMs in traditional NLP tasks, such as Named Entity Recognition. Recent benchmarks such as HealthSearchQA{~\citep{singhal2023large}} and WebMedQA{~\citep{he2019applying}} have shifted to open-ended formats, testing LLMs' ability to generate coherent responses for patient inquiries. 
While the above benchmarks evaluate factual accuracy and response quality, there's a gap in assessing LLMs for doctor-specific tasks.

\textbf{Medical Instruction and Conversation  Data}
Several datasets such as Huatuo-26M{~\citep{li2023huatuo}}, HealthCareMagic100k{~\citep{yunxiang2023chatdoctor}}, ChatDoctor10k{~\citep{yunxiang2023chatdoctor}}, and MedDialog{~\citep{zeng2020meddialog}} focus on online patient-doctor dialogues, primarily addressing outpatient inquiries without extending to inpatient scenarios. Additionally, structured resources like KG and MCQAs, including cMedQA2{~\citep{zhang2018multi}} and MedMCQA{~\citep{pal2022medmcqa}}, offer structured medical knowledge but fall short in providing the depth needed for comprehensive clinical responses.

\section{Conclusion}
In this paper, we emphasize a paradigm shift from developing \colorbox{as}{\strut LLMs as Doctors} to \colorbox{for}{\strut LLMs for Doctors}. We conduct a two-stage survey to accurately determine the real-world needs of doctors and create the DoctorFLAN dataset to address the training deficiencies of previous models. Additionally, we introduce DotaBench as a complementary evaluation framework to evaluate the ability of popular medical LLM as the doctor assistant. The benchmarking results show that most existing LLM models still face difficulties in being competent as medical assistants, but the superiority of DotaGPT demonstrates that the dataset we construct can help improve this ability. It proves that the dataset and benchmark we construct are effective and a beneficial supplement to current medical LLM research.

\newpage

\section*{Limitations}
The DoctorFLAN is currently only available in Chinese and may require supplementation in other languages. Consequently, it cannot be guaranteed that DotaGPT trained on DoctorFLAN will perform well in languages other than the one it has been tested. However, the methodology employed to create DoctorFLAN can be applied universally across different languages. Additionally, although DotaGPT has demonstrated impressive performance on the benchmarks, it is important to exercise caution when using its outputs, particularly in real-world doctor-assistant interactions.

\section*{Ethics Statement}
Ensuring the privacy and security of data is paramount in the development of medical applications. The datasets used in this study, primarily derived from Medtiku, an open-source repository of medical examination questions, and PromptCBLUE\cite{zhu2023promptcblue}, are both freely available for use. Additionally, we incorporate data from an internet medical encyclopedia hosted by 120 Ask\footnote{https://m.120ask.com/}, also open to the public.

Given the potential issues with the credibility of content generated by DotaGPT, we are committed to strictly regulating the model's use to prevent misuse. Our datasets, DoctorFLAN and DotaBench, will be released under terms that uphold the highest ethical standards. This commitment ensures that while advancing the capabilities of large language models in healthcare, we also safeguard sensitive medical data.

\bibliography{custom}

\appendix
\section{Dataset Comparison}
\label{appendix:data_comparison}
\begin{figure}
    \centering
    \includegraphics[width=\columnwidth]{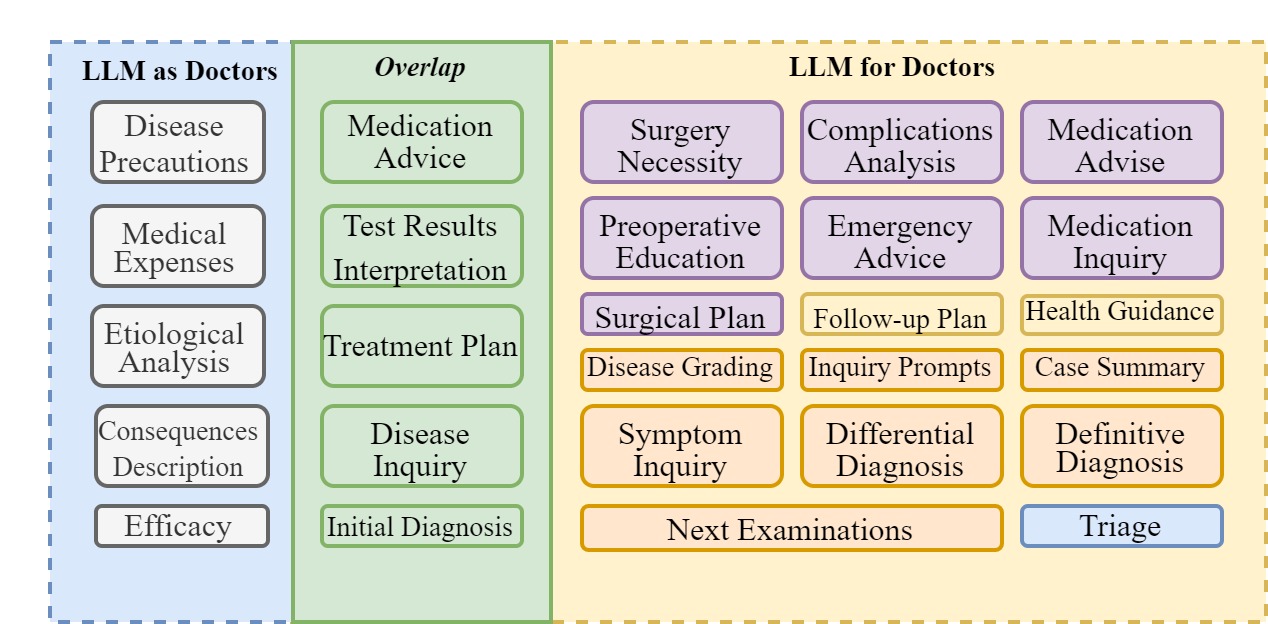}
    \caption{Visual comparison of task overlap between \colorbox{as}{\strut LLMs as Doctors} and \colorbox{for}{\strut LLMs for Doctors} datasets, illustrating unique and shared tasks in DoctorFLAN.}
    \label{fig:task_overlap}
\end{figure}

We compare tasks within two types of datasets, using KUAKE-QIC\cite{zhang2021cblue}\footnote{An evaluation dataset used by HuatuoGPT\cite{zhang2023huatuogpt}.} as a representative example of \colorbox{as}{\strut LLMs as Doctors} datasets and DoctorFLAN for \colorbox{for}{\strut LLMs for Doctors} datasets. We find while there is some overlap in tasks between these datasets, DoctorFLAN includes 17 additional tasks not present in KUA KE-QIC, as detailed in Figure \ref{fig:task_overlap}. These additional tasks reflect DoctorFLAN's broader coverage of in-hospital medical scenarios, such as surgery planning, which are generally absent in online consultation contexts.

\section{Dataset and Corresponding URLs}
\label{appendix:urls}
Below are the datasets used in our research along with their corresponding URLs:
\begin{itemize}
    \item HealthCareMagic100k: \url{https://github.com/Kent0n-Li/ChatDoctor}
    \item ChatDoctor10k: \url{https://github.com/Kent0n-Li/ChatDoctor}
    \item MedDialog: \url{https://github.com/UCSD-AI4H/Medical-Dialogue-System}
    \item CMeKG: \url{https://github.com/king-yyf/CMeKG}
    \item cMedQ\&A2: \url{https://github.com/zhangsheng93/cMedQ\&A2}
    \item MedMCQ\&A: \url{https://medmcqa.github.io}
\end{itemize}

\section{Details of Healthcare Professionals Involved in the Symposium}
\label{appendix:details_of_symposium}
The symposium convenes healthcare professionals from four distinct medical specialties: internal medicine, gynecology, surgery, and pediatrics. Each specialty is represented by four experts, comprising two mid-level and two senior specialists. 

\section{Details of the Survey Process}

\label{appendix:survey-questionnaire}
In this section, we present the distribution process of the questionnaire, our method for selecting valid responses, and the full text of the questionnaire itself.
\subsection{Design of the Survey}
The questionnaire is divided into two main sections. The first section aims to gauge doctors' evaluations of the usefulness of 22 specific tasks to their work efficiency, rated on a scale of 1 to 5. A score of 4 signifies that the LLM assistance is helpful but not essential, while a score of 5 indicates its crucial role in significantly enhancing efficiency. The second section seeks to understand doctors' preferences regarding the LLMs.
\subsection{Survey Target Respondents}
Doctors are sampled from over 12 Grade A tertiary care hospitals to ensure the universality of the tasks we developed. Feedback and suggestions are gathered from these doctors through the questionnaires.
\subsection{Survey Collection}
We collect a total of 82 responses, which are then refined to 71 valid responses through a detailed review. The validity of the responses is determined based on two stringent criteria: 1) the completion time is less than one-third of the average duration observed across all submissions, indicating a potential lack of due consideration; and 2) a marked uniformity in responses (e.g., repetitive selection of the same response option), suggesting insufficient engagement with the content of the questionnaire. This dual-condition criterion ensures that respondents have engaged with the material thoughtfully, recognizing and reflecting on the varied tasks the survey examined. On average, respondents spend 191.82 seconds on the questionnaires deemed valid. Importantly, the survey is designed to protect participant anonymity by not collecting any personal information.

\subsection{Detailed Questionnaire on Medical AI Assistant Requirements}
Below is the English translation of the original questionnaire, which is provided to the doctors in Chinese.\\
\textbf{Dear Doctor,}

In your daily medical practice, if there is a medical AI assistant ("AI Assistant") available to provide assistance, which problems would you most like it to solve? To better meet your needs, we have listed tasks that may improve medical efficiency and envisioned how the AI assistant could assist with these tasks. You are invited to assess the importance of these tasks according to the scoring guide.

\textbf{Scoring Guide:}
\begin{itemize}
\item Very Important (5 points): Indicates that the assistance of the AI assistant will significantly improve your work efficiency (High Efficiency Score).
\item Important (4 points): Indicates that the AI assistant can improve your work efficiency, but it is not crucial (Moderate Efficiency Score).
\item Neutral (3 points): Indicates that the AI assistant provides some assistance to your work, but it is not essential (Neutral Efficiency Score).
\item Not Very Important (2 points): Indicates that the AI assistant will not be of much help to your work (Low Efficiency Score).
\item Not Important At All (1 point): Indicates that the assistance of the AI assistant in the task is completely unhelpful to your work (No Efficiency Score).
\end{itemize}

\subsection*{Pre-diagnosis}
\begin{enumerate}
    \item Guide to Diagnosis: Based on the patient's condition, the medical assistant directs the patient to the correct department. [Single choice]*
    \begin{itemize}
        \item Not Important At All
        \item Not Very Important
        \item General
        \item Important
        \item Very Important
    \end{itemize}
\end{enumerate}

\subsection*{During Diagnosis}
\begin{enumerate}[resume]
    \item Inquiry prompt: During the doctor-patient conversation, the medical assistant provides prompts and suggestions for inquiries. [Single choice]*
    \begin{itemize}
        \item Not Important At All
        \item Not Very Important
        \item General
        \item Important
        \item Very Important
    \end{itemize}
    
    \item Symptom query: You query information related to symptoms (such as causes of the symptom) based on symptom description, and the medical assistant returns the query results. [Single choice]*
    \begin{itemize}
        \item Not Important At All
        \item Not Very Important
        \item General
        \item Important
        \item Very Important
    \end{itemize}
    
    \item Disease query: You query disease-related information (such as specific disease diagnosis process and methods, complications, and prognosis) based on the disease name, and the medical assistant returns the query results. [Single choice]*
    \begin{itemize}
        \item Not Important At All
        \item Not Very Important
        \item General
        \item Important
        \item Very Important
    \end{itemize}
    
    \item Initial diagnosis: The medical assistant assists the doctor in preliminary diagnosis decisions and analysis of the condition. [Single choice]*
    \begin{itemize}
        \item Not Important At All
        \item Not Very Important
        \item General
        \item Important
        \item Very Important
    \end{itemize}
    
    \item Summarizing the medical record from the initial consultation dialogue: The medical assistant automatically summarizes the medical record information based on the content of the doctor-patient conversation. [Single choice]*
    \begin{itemize}
        \item Not Important At All
        \item Not Very Important
        \item General
        \item Important
        \item Very Important
    \end{itemize}
    
    \item Differential diagnosis: The medical assistant provides differential diagnosis assistance based on the patient's diagnostic information. [Single choice]*
    \begin{itemize}
        \item Not Important At All
        \item Not Very Important
        \item General
        \item Important
        \item Very Important
    \end{itemize}
    
    \item Next steps for examination: After providing the patient's information to the medical assistant, it can suggest further examination and testing recommendations for the patient. [Single choice]*
    \begin{itemize}
        \item Not Important At All
        \item Not Very Important
        \item General
        \item Important
        \item Very Important
    \end{itemize}
    
    \item Interpretation of examination results: The medical assistant assists in interpreting the results of examinations and tests. [Single choice]*
    \begin{itemize}
        \item Not Important At All
        \item Not Very Important
        \item General
        \item Important
        \item Very Important
    \end{itemize}
    
    \item Confirmation of diagnosis: Based on all the patient's diagnostic information, the medical assistant can assist the doctor in making the final diagnosis. [Single choice]*
    \begin{itemize}
        \item Not Important At All
        \item Not Very Important
        \item General
        \item Important
        \item Very Important
    \end{itemize}
    
    \item Disease classification and grading: The medical assistant classifies and grades the disease based on the severity of the patient's condition. [Single choice]*
    \begin{itemize}
        \item Not Important At All
        \item Not Very Important
        \item General
        \item Important
        \item Very Important
    \end{itemize}
\end{enumerate}

\subsection*{Treatment Phase}
\begin{enumerate}[resume]
    \item Emergency care: In case of an emergency situation with the patient, the medical assistant provides guidance and suggestions for emergency treatment measures. [Single choice]*
    \begin{itemize}
        \item Not Important At All
        \item Not Very Important
        \item General
        \item Important
        \item Very Important
    \end{itemize}
    
    \item Treatment plan: Based on the patient's medical record, the medical assistant provides possible treatment plan suggestions. [Single choice]*
    \begin{itemize}
        \item Not Important At All
        \item Not Very Important
        \item General
        \item Important
        \item Very Important
    \end{itemize}
    
    \item Drug query: You query information related to a drug (such as contraindications, etc.) based on the drug name, and the medical assistant returns the query results. [Single choice]*
    \begin{itemize}
        \item Not Important At All
        \item Not Very Important
        \item General
        \item Important
        \item Very Important
    \end{itemize}
    
    \item Medication advice: Based on the patient's medical record, the medical assistant provides medication advice and plans. [Single choice]*
    \begin{itemize}
        \item Not Important At All
        \item Not Very Important
        \item General
        \item Important
        \item Very Important
    \end{itemize}
    
    \item Complications: Based on the patient's medical record, the medical assistant helps identify and prevent possible complications. [Single choice]*
    \begin{itemize}
        \item Not Important At All
        \item Not Very Important
        \item General
        \item Important
        \item Very Important
    \end{itemize}
    
    \item Adjustment of the treatment plan: The medical assistant will suggest possible adjustments to the treatment plan based on changes in the patient's condition recorded in the medical record. [Single choice]*
    \begin{itemize}
        \item Not Important At All
        \item Not Very Important
        \item General
        \item Important
        \item Very Important
    \end{itemize}
    
    \item Necessity of surgery: Based on the patient's medical record, the medical assistant assesses whether the patient indeed needs to undergo surgery. [Single choice]*
    \begin{itemize}
        \item Not Important At All
        \item Not Very Important
        \item General
        \item Important
        \item Very Important
    \end{itemize}
    
    \item Surgery plan: For patients needing surgery, the medical assistant provides surgery plans and related suggestions. [Single choice]*
    \begin{itemize}
        \item Not Important At All
        \item Not Very Important
        \item General
        \item Important
        \item Very Important
    \end{itemize}
    
    \item Pre-surgery education: The medical assistant provides pre-surgery education and preparation information for patients about to undergo surgery. [Single choice]*
    \begin{itemize}
        \item Not Important At All
        \item Not Very Important
        \item General
        \item Important
        \item Very Important
    \end{itemize}
\end{enumerate}

\subsection*{Post-diagnosis}
\begin{enumerate}[resume]
    \item Health guidance: Based on the patient's medical record, the medical assistant provides guidance on a healthy lifestyle. [Single choice]*
    \begin{itemize}
        \item Not Important At All
        \item Not Very Important
        \item General
        \item Important
        \item Very Important
    \end{itemize}
    
    \item Follow-up plan: Based on the patient's medical record, the medical assistant assists in arranging the patient's subsequent follow-up plan. [Single choice]*
    \begin{itemize}
        \item Not Important At All
        \item Not Very Important
        \item General
        \item Important
        \item Very Important
    \end{itemize}
\end{enumerate}

\subsection*{Other Questions}
\begin{enumerate}[resume]
    \item Beyond the tasks listed in the four stages above, other tasks you think require the assistance of a medical assistant (optional). [Open-ended question]
    
    \item As a doctor, which type of disease-related information are you most concerned about? [Single choice]*
    \begin{itemize}
        \item A Detailed physiological and pathological mechanisms of the disease
        \item B Epidemiological data of the disease
        \item C Latest treatment advancements and clinical guidelines related to the disease
        \item D Diagnostic methods of the disease
        \item E Other \_\_\_\_\_\_\_\_\_
    \end{itemize}
    
    \item As a doctor, which type of drug-related information are you most concerned about? [Single choice]*
    \begin{itemize}
        \item A Mechanism of action of the drug
        \item B Indications and usage of the drug
        \item C Interactions and side effects of the drug
        \item D Safety of the drug and precautions to be taken
        \item E Other \_\_\_\_\_\_\_\_\_
    \end{itemize}
    
    \item As a doctor, when you query information related to treatment plans, which aspects do you usually pay most attention to? [Single choice]*
    \begin{itemize}
        \item A Detailed treatment methods and means
        \item B Cost and effectiveness of the treatment plan
        \item C Adjustments to lifestyle and precautions during the treatment period
        \item D Potential complications and side effects of the treatment plan
        \item E Other \_\_\_\_\_\_\_\_\_
    \end{itemize}
    
    \item As a doctor, you prefer the answers from the medical assistant to be [Single choice]*
    \begin{itemize}
        \item A Concise and easy to understand
        \item B Professional and clinically accurate
        \item C Detailed explanation
        \item D Friendly and gentle in tone
    \end{itemize}
    
    \item Have you ever used large language models (e.g., GPT-3.5/GPT-4/Wenxin Yiyuan/Huatuo) in your work? [Single choice]*
    \begin{itemize}
        \item Yes
        \item No (Please skip to the end of the questionnaire and submit your answers)
    \end{itemize}
    
    \item The most frequently used large language model is [Single choice]*
    \begin{itemize}
        \item GPT-3.5
        \item GPT-4
        \item Wenxin Yiyan
        \item HuatuoGPT
        \item Other \_\_\_\_\_\_\_\_\_
    \end{itemize}
    
    \item Please rate your satisfaction with using this model. [Single choice]*
    \begin{itemize}
        \item Very dissatisfied
        \item Dissatisfied
        \item Neutral
        \item Satisfied
        \item Very satisfied
    \end{itemize}
    
    \item Reasons for dissatisfaction [Multiple choice]*
    \begin{itemize}
        \item[$\square$] A Often unable to follow my commands or understand my questions
        \item[$\square$] B The large language model often cannot correctly answer my questions
        \item[$\square$] C The large language model cannot provide correct justifications, and I cannot trust its conclusions
        \item[$\square$] D Other \_\_\_\_\_\_\_\_\_
    \end{itemize}
\end{enumerate}


\section{Data Statistics}
\label{appendix:data statistics}
Table \ref{tab: data volume} provides a detailed breakdown of the data volume corresponding to each specific task within the DoctorFLAN.
Table \ref{tab:category} showcases the distribution of cases as a percentage across different specialties.
\begin{table*}[ht]
\resizebox{\linewidth}{!}{%
\begin{tabular}{lllll}
\hline
Phase & Key Task & Task Description & n & Original Data Source\\ \hline
Pre-Diagnosis & Triage & Provide department suggestions based on patient's condition & 2535 & MCQA\\ \hline
\multirow{11}{*}{During Diagnosis} & Inquiry Prompts & Guide further questioning based on patient's medical history & 1000 & MCQA \\
& Symptom Inquiry & Retrieve encyclopedic knowledge related to the symptom & 6000 & Encyclopedia \\ 
& Disease Inquiry & Retrieve encyclopedic knowledge related to the disease & 12000 & Encyclopedia\\
& Initial Diagnosis & Identifying potential diseases based on patient history and simple tests & 13358  & MCQA\\
& Case Summary & Compile key points from doctor-patient dialogue into a patient case & 2807 & PromptCBLUE\\
& Differential Diagnosis & Distinguish between diseases with similar symptoms & 1000  & MCQA\\
& Next Examinations & Decide on necessary laboratory or imaging tests for diagnosis confirmation & 6500 & MCQA\\
& Test Results Interpretation & Explain the meaning of patient's test results & 1000  & MCQA\\
& Definitive Diagnosis & Confirm the final diagnosis based on all available information & 15062  & MCQA\\
& Disease Grading & Determine the category and stage of the disease based on established medical criteria & 1000  & MCQA\\
\hline
\multirow{9}{*}{Treatment} & Emergency Advice & Provide guidance on handling emergencies and surgical situations & 1043  & MCQA\\
& Treatment Plan & Develop a suitable treatment plan based on the diagnosis & 5923  & MCQA\\
& Medication Inquiry & Retrieve encyclopedic knowledge related to the medicine & 10000  & Encyclopedia\\
& Medication Advice & Provide specific medication recommendations for the patient & 2621 & MCQA \\
& Complications Analysis & Identify potential and preventable complications & 1000 & MCQA\\
& Treatment Adjustment & Modify the treatment plan based on the patient's response to initial treatment & 1000  & MCQA\\
& Surgery Necessity & Evaluate if surgical treatment is necessary and beneficial & 4708 & MCQA\\
& Surgical Plan & Develop a detailed surgical plan based on the patient's condition & 1000 & MCQA\\ 
& Preoperative Education & Educate the patient about the surgical procedure and postoperative care & 769 & MCQA\\ \hline
\multirow{2}{*}{Post-Treatment} & Health Guidance & Provide advice on maintaining health and preventing disease recurrence & 1000 & MCQA\\
& Follow-up Plan & Develop a plan for regular check-ups and ongoing care & 1000 & MCQA\\ \hline
Total &  &  & 91,800 \\
\hline
\end{tabular}%
}
\caption{Data Volume Distribution for Various Tasks in the DoctorFLAN. MCQA represents multiple-choice Question Answer.}
\label{tab: data volume}
\end{table*}

\begin{table}[ht]
\footnotesize
\centering
\begin{tabular}{l|l}
\hline
\textbf{Specialist} & \textbf{Percentage (\%)} \\
\hline
Gastroenterology & 7.81 \\
Pediatrics & 7.25 \\
Obstetrics \& Gynecology & 6.74 \\
Respiratory Medicine & 6.83 \\
Cardiology & 6.65 \\
Neurology & 6.34 \\
General Surgery & 5.73 \\
Stomatology & 5.01 \\
Nephrology & 4.72 \\
Hepatology & 4.06 \\
Orthopedics & 3.70 \\
Urology & 3.78 \\
Hematology & 3.82 \\
Endocrinology & 3.76 \\
Oncology & 3.64 \\
Emergency Medicine & 3.14 \\
Infectious Disease & 3.13 \\
Traditional Chinese Medicine & 2.76 \\
Ophthalmology & 1.86 \\
Rheumatology \& Immunology & 1.75 \\
Neurosurgery & 1.78 \\
Dermatology & 1.73 \\
Otorhinolaryngology(ENT) & 1.50 \\
Vascular Surgery & 0.74 \\
Multidisciplinary & 0.62 \\
Spine Surgery & 0.55 \\
Cardiothoracic Surgery & 0.31 \\
Other & 0.30 \\
\hline
\end{tabular}
\caption{Percentage of cases per specialist}
\label{tab:category}
\end{table}

\section{Verification Details of DoctorFLAN Data}
\label{appendix:DoctorFLAN_checking}

\subsection{Data Verification Setting}
Due to the extensive size of the dataset and budget constraints, full verification is not feasible; instead, we sample 50 data points per task, resulting in a total of 1050 samples for doctors to review. A senior professional with a high-level title conducted the verification, dedicating 10 hours to the task. This approach balances thoroughness with practical limitations, ensuring credible verification within the available resources.

To ensure the reliability of our dataset, we establish specific criteria for data verification focusing on the following aspects:

\textbf{Correctness:} This criterion evaluates the accuracy of the "GPT-4 Answer". An answer is deemed correct if it is factually accurate within the medical context. If the "GPT-4 Answer" is entirely correct and free of medical factual errors, please select "Yes"; otherwise, select "No".

\textbf{Practicality:} This criterion assesses the usefulness of the "GPT-4 Answer" compared to the "Original Answer". If the "GPT-4 Answer" is deemed more practical, select "Yes"; if not, select "No".

\begin{figure*}[t]
\centering
\includegraphics[width=0.8\textwidth]{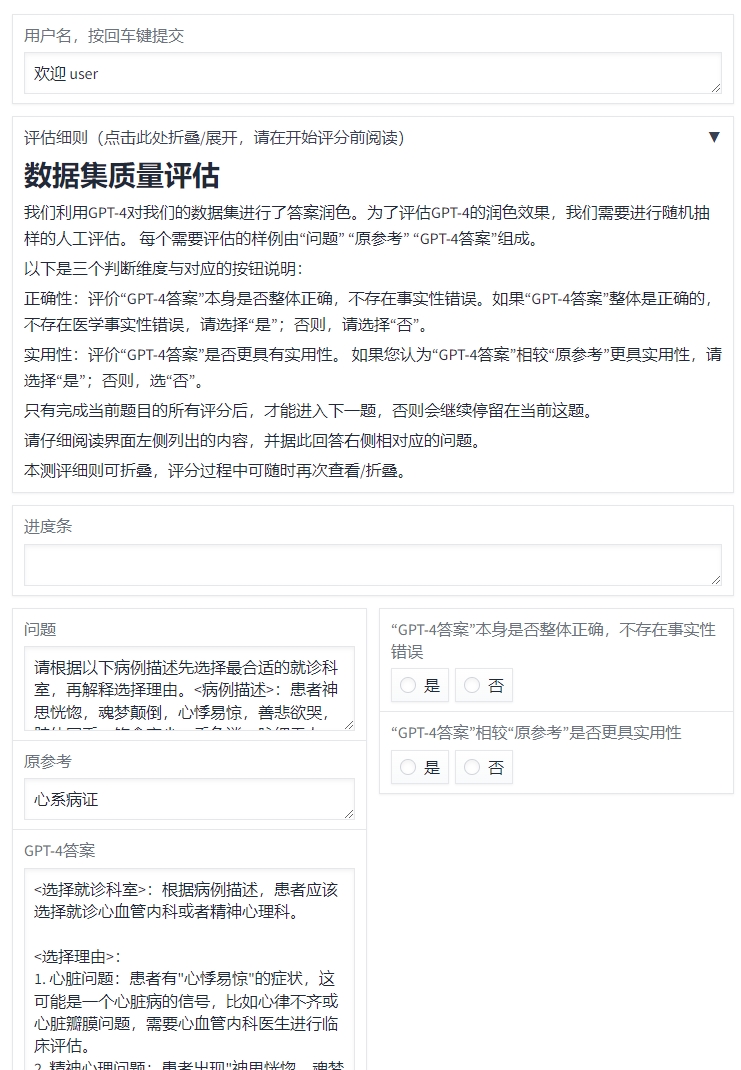}
\caption{Interface for Data Verification}
\label{fig: gradio_data_verification}
\end{figure*}

\begin{figure*}[t]
\centering
\includegraphics[width=0.8\textwidth]{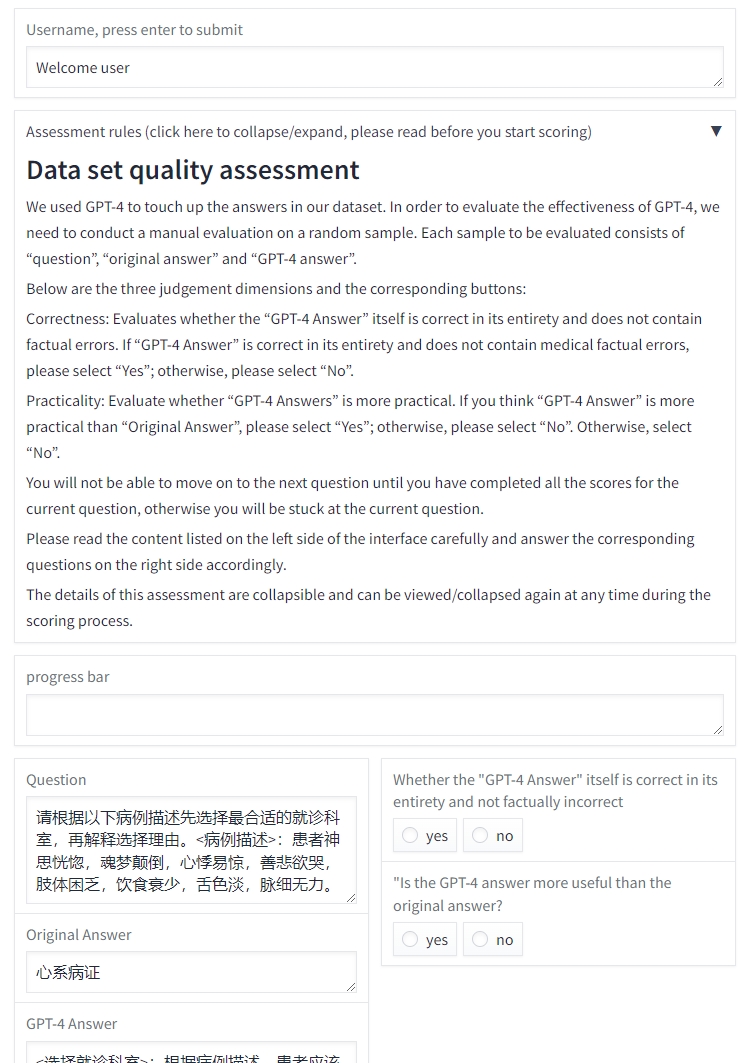}
\caption{Interface for Data Verification (Translated)}
\label{fig: gradio_data_verification_en}
\end{figure*}

Illustrations of the graphical user interface used for this verification are shown in Figures \ref{fig: gradio_data_verification} and \ref{fig: gradio_data_verification_en}.

\subsection{Case Study: A Failure in Data Verification}
In a detailed examination of the data verification stage, we identify an instance where a doctor noted the lack of practicality, commenting on the "lack of specific details.", as shown in Figure \ref{fig:verification failure}. Such feedback suggests that the responses refined by GPT-4 can sometimes fall short in complex practical medical contexts, highlighting an area for future improvement.
\begin{figure*}[t]
\begin{AIbox}{An Example Lacking Practicality Compared to the Original Answer}
{\bf Instruction:} \\
{\footnotesize
脸部疼痛:提供症状的详细概述、可能的原因以及与之相关的疾病。包括症状的定义、典型表现、可能引起症状的常见病因，以及与该症状相关的其他健康问题。\\
{\it Translation:} Facial pain: Provide a detailed overview of the symptoms, possible causes, and related diseases. Include the definition of the symptoms, typical manifestations, common causes that may cause the symptoms, and other health issues related to these symptoms.\\
}\\
{\bf Reference:} \\
{\footnotesize
associated\_diseases:颞下间隙感染、慢性上颌窦炎、颞动脉炎、下颌下间隙感染、颌下间隙感染、颌骨骨髓炎、骨囊肿、颞骨岩部炎、急性颌骨骨髓炎、急性颌下腺炎\\
overview:脸部疼痛是指在脸部的一边或者双边疼痛，或者额头处疼痛。这种疼痛可以是钝痛，也可以是波动痛，还可能是非常强烈的刺痛。面部疼痛或痉挛，一般没有生命攸关的威胁，但痛者却十分痛苦，即使能忍住疼痛，表情麻木不自然。\\
cause:面部疼痛通常由局部组织感染或炎症引起，有的是隐隐的跳痛，有的是剧烈的刺痛，表现出的病症不太一样。肌肉发炎是最为人所熟悉的脸痛原因，也叫颞颌关节异常。其次三叉神经炎，带状疱疹。\\
也可以见于口腔炎症刺激导致的面部疼痛，也可以见于局部外伤刺激导致的疼痛情况的。\\
{\it Translation:} associated\_diseases: Infection in the infratemporal space, chronic maxillary sinusitis, temporal arteritis, infection under the lower jaw, infection in the submandibular space, osteomyelitis of the jaw, bone cyst, petrositis of the temporal bone, acute osteomyelitis of the jaw, acute submandibular sialadenitis\\
overview: Facial pain refers to pain on one or both sides of the face, or at the forehead. This pain can be dull or throbbing and can be extremely sharp. Facial pain or spasms, generally not life-threatening, but the sufferer is very miserable; even if the pain is bearable, the facial expression appears unnatural.\\
cause: Facial pain is usually caused by local tissue infection or inflammation; some are throbbing pain, some are severe stabbing pain, and the presented symptoms vary. Muscle inflammation is the most familiar cause of facial pain, also known as temporomandibular joint disorder. Followed by trigeminal neuralgia, herpes zoster.\\
It can also be seen in facial pain caused by oral inflammation, and pain caused by local trauma.\\
}\\
{\bf Answer:} \\
{\footnotesize
面部疼痛是临床上比较常见的症状，其特点和表现形式多种多样。患者可能会描述面部疼痛为钝痛、波动痛、烧灼感或者极端的刺痛。在考虑面部疼痛的病因时，需要综合考虑痛觉的分布、性质、诱发和缓解因素以及伴随症状。\\
根据<书本知识>，与面部疼痛相关联的疾病包括但不限于颞下间隙感染、慢性上颌窦炎、颞动脉炎、下颌下间隙感染、颌下间隙感染、颌骨 \\
{\it Translation:} Facial pain is a relatively common clinical symptom, characterized by various forms and manifestations. Patients may describe facial pain as dull pain, throbbing pain, burning sensation, or extreme stabbing pain. When considering the causes of facial pain, it is necessary to comprehensively consider the distribution, nature, triggering and alleviating factors, and accompanying symptoms.\\
According to <textbook knowledge>, diseases associated with facial pain include, but are not limited to, infection in the infratemporal space, chronic maxillary sinusitis, temporal arteritis, infection under the lower jaw, infection in the submandibular space, and osteomyelitis of the jaw.
}\\
{\bf Doctor Comment:} \\
{\footnotesize
没有原参考答案具体详细\\
{\it Translation:} The answer lacks specific detail compared to the original reference answer.
}
\end{AIbox}
\caption{Example of a response judged by doctors to be less practical compared to the original, illustrating a potential challenge in ensuring the utility of LLM-generated content in medical data.}
\label{fig:verification failure}
\end{figure*}



\section{Model Details}
\label{appendix:model details}
\label{appendix:details of existing models}

\textbf{BianQue-2} 
BianQue-2 is a medical language model fine-tuned on ChatGLM-6B \citep{zeng2022glm}, specifically oriented towards patient-centric conversations. 

\textbf{DISC-MedLLM}. DISC-MedLLM is a domain-specific model crafted for conversational healthcare, leveraging the high-quality DISC-Med-SFT dataset. It is built upon the Baichuan-13B-Base architecture and is tailored for deep interactions within the medical domain.

\textbf{HuatuoGPT}. An advanced language model specifically fine-tuned from the Baichuan-7B architecture for Chinese medical consultations. HuatuoGPT is trained on a comprehensive Chinese medical corpus with the goal to provide professional-grade support in medical consultation scenarios, enhancing the decision-making process in healthcare.

\textbf{HuatuoGPT-II}. Recognized as the state-of-the-art among Chinese medical language models, HuatuoGPT-II utilizes the Baichuan2-7B-Base architecture. The 7B version, chosen for our evaluations, integrates extensive medical knowledge.

\textbf{Qwen-1.8B-Chat}
Part of the Qwen series, this general-purpose language model is trained on trillions of tokens. It employs advanced techniques such as Supervised Fine-Tuning (SFT) and Reinforcement Learning from Human Feedback (RLHF) to enhance its performance across diverse conversational contexts.

\textbf{Baichuan-13B-Chat~\footnote{https://huggingface.co/baichuan-inc/Baichuan-13B-Chat}}. An extension of the Baichuan-13B-Base, this model integrates enhanced dialogue capabilities in both Chinese and English. 

\textbf{Baichuan2-7B-Chat}. 
Derived from the Baichuan2-7B-Base, this model shares a common architecture with several other specialized models but is fine-tuned for more general applications, balancing robustness and versatility.

\textbf{Baichuan2-13B-Chat}. An extension of the Baichuan2-13B-Base, this model is enhanced to handle more complex and diverse tasks in conversational AI. It offers increased computational power and improved response quality, suitable for advanced conversational systems.

\textbf{Yi-6B-Chat}
Yi-6B-Chat excels in both English and Chinese dialogue, aligning closely with user preferences. While it performs well, it exhibits some limitations when compared to more advanced models like GPT-4.

\textbf{Yi-34B-Chat}
The larger counterpart within the Yi series, Yi-34B-Chat, showcases enhanced performance in bilingual settings and is favored in platforms such as AlpacaEval and Chatbot Arena. It outperforms the Yi-6B-Chat in handling complex dialogues and user interactions but still trails behind the capabilities of GPT-4.

\textbf{Claude-3}
Developed by Anthropic, Claude-3~\footnote{We use Claude 3 Opus.} is designed to engage in natural language conversations across a broad spectrum of tasks, emphasizing human-like text generation and comprehension.

\textbf{GPT-3.5}
A product of OpenAI, GPT-3.5~\footnote{We use gpt-3.5-turbo-0125} is designed to deliver conversational interactions, drawing on a vast corpus of diverse textual data to generate responsive and contextually appropriate dialogues.

\textbf{GPT-4}
As the successor to GPT-3.5, GPT-4~\footnote{We use gpt-4-turbo-2024-04-09.} by OpenAI advances further in language comprehension and generation, significantly enhancing interaction quality and precision in diverse conversational contexts.

\section{Details of Model Training}
\label{appendix:model_training}
To prevent catastrophic forgetting of general knowledge, we use general datasets like Evol-instruct{~\citep{xu2023wizardlm}}, ShareGPT~\footnote{https://sharegpt.com/} in addition to DoctorFLAN-\textit{training} and supplemented it. Furthermore, to augment the model's proficiency in medical knowledge, we have included data from the training set of CMExam~\citep{liu2023benchmarking} in our training regimen. We train our model with a maximum sequence length of 4096 to accommodate conversational data. The training parameters included a learning rate of 5e-5 and a batch size of 4 per GPU, conducted 3 epochs on four A100 GPUs.

\begin{table*}[t]
\centering
\resizebox{1\linewidth}{!}{%
\setlength{\tabcolsep}{3pt}
\begin{tabular}{lccccccccccccccc}
\hline

\multirow{2}{*}{\textbf{Task}} & \multirow{2}{*}{\textbf{Qwen-1.8B-Chat}} &  \multirow{2}{*}{\textbf{Baichuan-13B-Chat}} & \multirow{2}{*}{\textbf{Baichuan2-7B-Chat}} & \multirow{2}{*}{\textbf{Baichuan2-13B-Chat}} & \multirow{2}{*}{\textbf{Yi-6B-Chat}} & \multirow{2}{*}{\textbf{Yi-34B-Chat}}  & \multirow{2}{*}{\textbf{BianQue-2}} & \multirow{2}{*}{\textbf{DISC-MedLLM}} & \multirow{2}{*}{\textbf{HuatuoGPT}} & \multirow{2}{*}{\textbf{HuatuoGPT-II}} & \textbf{DotaGPT} & \textbf{DotaGPT} & \multirow{2}{*}{\textbf{GPT-3.5}} & \multirow{2}{*}{\textbf{Claude-3}} & \multirow{2}{*}{\textbf{GPT-4}}\\
&&&&&&&&&&&\textbf{(Baichuan2-7B-Base)} & \textbf{(Yi-6B)} & &&\\
\hline
Triage & 5.28 & 6.20 & 6.32 & 6.76 & 7.00 & 7.36 & 5.56 & 5.56 & 5.32 & 7.60 & \textbf{8.48} & \underline{8.32} & 6.40 & 7.80 & 8.00 \\
Inquiry Prompts & 7.16 & 8.39 & 8.08 & 8.49 & 8.92 & 8.20 & 2.00 & 6.56 & 4.48 & 8.20 & \underline{9.04} & 9.04 & 8.12 & 8.96 & \textbf{9.12} \\
Symptom Inquiry & 5.80 & 6.68 & 6.56 & 7.13 & 7.24 & 7.56 & 2.80 & 3.76 & 4.72 & 7.44 & 8.20 & 7.24 & 6.76 & \underline{8.32} & \textbf{8.52} \\
Disease Inquiry & 3.40 & 6.92 & 6.44 & 7.51 & 6.44 & 7.44 & 2.76 & 4.64 & 4.56 & 7.28 & 7.52 & 6.96 & 6.64 & \textbf{8.4} & \underline{8.04} \\
Initial Diagnosis & 4.44 & 5.47 & 5.60 & 6.00 & 6.44 & 6.84 & 3.96 & 4.80 & 4.72 & 7.36 & 7.80 & \textbf{8.44} & 6.08 & \underline{8.36} & 8.16 \\
Case Summary & 5.64 & 8.52 & 8.84 & 8.93 & 8.28 & 8.56 & 1.28 & 3.20 & 4.48 & 7.04 & 6.96 & 6.48 & 9.08 & \textbf{9.32} & \underline{9.24} \\
Differential Diagnosis & 4.04 & 6.41 & 6.64 & 7.09 & 6.72 & 6.64 & 4.28 & 4.52 & 4.28 & 7.00 & \underline{8.68} & 7.56 & 6.64 & 8.44 & \textbf{8.88} \\
Next Examinations & 5.52 & 7.45 & 7.24 & 7.88 & 8.00 & 7.88 & 5.68 & 4.60 & 4.48 & 6.84 & 8.80 & 8.32 & 7.12 & \textbf{8.92} & \underline{8.88} \\
Test Results Interpretation & 3.80 & 6.44 & 6.64 & 6.63 & 7.12 & 8.36 & 4.24 & 3.92 & 4.28 & 7.28 & \underline{8.72} & 8.20 & 6.64 & 8.72 & \textbf{8.84} \\
Definitive Diagnosis & 2.96 & 4.15 & 3.84 & 5.03 & 4.52 & 6.40 & 2.92 & 3.52 & 3.36 & 6.68 & \textbf{7.56} & \underline{6.96} & 6.20 & 6.72 & 6.56 \\
Disease Grading & 2.88 & 4.68 & 3.76 & 3.85 & 4.64 & 5.96 & 2.80 & 2.80 & 3.04 & 5.12 & 6.84 & 7.04 & 5.20 & \underline{7.64} & \textbf{7.84} \\
Emergency Advice & 3.76 & 5.81 & 7.60 & 7.52 & 7.08 & 8.68 & 3.28 & 3.28 & 3.24 & 7.40 & \textbf{8.88} & 8.56 & 6.32 & 8.84 & \underline{8.88} \\
Treatment Plan & 4.48 & 7.04 & 6.56 & 7.45 & 7.20 & 8.56 & 3.92 & 4.12 & 3.88 & 6.72 & \textbf{8.92} & 8.44 & 7.28 & 8.54 & \underline{8.88} \\
Medication Inquiry & 3.56 & 6.67 & 5.60 & 6.56 & 6.96 & 5.88 & 3.08 & 3.96 & 4.60 & 6.88 & \textbf{8.00} & \underline{7.48} & 6.32 & 7.36 & 6.60 \\
Medication Advice & 2.48 & 4.07 & 4.04 & 4.87 & 4.88 & 7.00 & 2.96 & 2.84 & 3.64 & 5.32 & 6.68 & 5.96 & 4.80 & \textbf{7.12} & \underline{7.08} \\
Complications Analysis & 4.04 & 6.61 & 6.28 & 7.11 & 7.84 & 8.28 & 4.28 & 3.20 & 3.48 & 6.80 & \textbf{8.84} & 7.68 & 6.60 & \underline{8.72} & 8.72 \\
Treatment Adjustment & 4.32 & 6.43 & 6.12 & 6.87 & 6.96 & 8.68 & 3.56 & 3.76 & 3.16 & 6.64 & 8.36 & 8.16 & 6.36 & \underline{8.77} & \textbf{8.84} \\
Surgery Necessity & 5.44 & 7.28 & 7.72 & 8.09 & 7.60 & 8.44 & 4.36 & 3.96 & 4.40 & 7.32 & 8.44 & 7.52 & 7.20 & \underline{8.64} & \textbf{8.92} \\
Surgical Plan & 3.60 & 6.61 & 6.80 & 7.07 & 6.08 & 8.04 & 3.76 & 3.16 & 3.36 & 6.40 & 7.72 & 7.64 & 5.20 & \textbf{8.28} & \underline{8.28} \\
Preoperative Education & 5.00 & 7.67 & 7.44 & 8.49 & 8.28 & \textbf{9.28} & 1.56 & 4.24 & 4.56 & 7.56 & \underline{9.04} & 8.84 & 7.60 & 9.00 & 8.88 \\
Health Guidance & 6.68 & 8.31 & 8.36 & 8.41 & 7.96 & \underline{9.00} & 5.36 & 5.72 & 5.64 & 7.92 & 9.00 & 8.60 & 7.68 & 8.92 & \textbf{9.08} \\
Follow-Up Plan & 4.20 & 6.80 & 7.04 & 7.20 & 7.36 & 8.56 & 4.20 & 4.56 & 4.20 & 6.92 & \underline{8.60} & 8.28 & 5.80 & 8.60 & \textbf{9.00} \\\hline
Average & 4.48 & 6.57 & 6.59 & 7.04 & 6.98 & 7.80 & 3.57 & 4.24 & 4.29 & 7.03 & 8.25 & 7.81 & 6.64 & \underline{8.38} & \textbf{8.42} \\
\hline
\end{tabular}
}
\caption{Detailed Auto Evaluation Results for the DoctorFLAN-\textit{test}}
\label{tab:new auto eval-S}
\end{table*}

\begin{table*}[t]
\centering
\resizebox{\linewidth}{!}{%
\setlength{\tabcolsep}{3pt}
\begin{tabular}{lcccccc}
\toprule
\textbf{Task} & \textbf{BianQue-2} & \textbf{HuatuoGPT} & \textbf{DISC-MedLLM} & \textbf{Baichuan2-7B-Chat} & \textbf{DotaGPT (Baichuan2-7B-Base)} & \textbf{GPT-4} \\
\hline
Triage & 4.24 & 5.04 & 5.00 & 6.28 & 7.00 & 7.76 \\
Inquiry Prompts & 3.72 & 4.28 & 6.00 & 5.88 & 7.88 & 7.68 \\
Symptom Inquiry & 3.60 & 7.08 & 6.00 & 8.60 & 8.68 & 8.84 \\
Disease Inquiry & 4.84 & 6.24 & 6.16 & 7.56 & 8.24 & 8.56 \\
Initial Diagnosis & 4.88 & 4.88 & 5.72 & 6.36 & 8.60 & 8.80 \\
Case Summary & 1.68 & 2.80 & 3.56 & 4.32 & 3.84 & 5.76 \\
Differential Diagnosis & 5.04 & 4.16 & 4.44 & 5.68 & 7.00 & 7.36 \\
Next Steps for Examination & 5.44 & 5.28 & 4.84 & 7.40 & 8.60 & 8.28 \\
Test Results Interpretation & 4.20 & 4.44 & 5.40 & 7.96 & 8.72 & 8.44 \\
Definitive Diagnosis & 4.96 & 4.32 & 4.84 & 4.96 & 8.36 & 7.84 \\
Disease Grading & 4.00 & 4.28 & 5.96 & 4.08 & 6.80 & 7.44 \\
Emergency Advice & 4.04 & 4.84 & 4.64 & 8.88 & 9.36 & 8.76 \\
Treatment Plan & 5.64 & 4.12 & 5.28 & 6.32 & 7.36 & 8.16 \\
Medication Inquiry & 5.72 & 5.24 & 5.76 & 6.48 & 8.68 & 7.60 \\
Complications Analysis & 5.04 & 4.24 & 5.00 & 6.20 & 7.32 & 8.20 \\
Treatment Adjustment & 5.36 & 4.96 & 6.04 & 7.32 & 7.72 & 8.56 \\
Surgery Necessity & 5.00 & 5.80 & 6.44 & 8.32 & 8.72 & 8.68 \\
Surgical Plan & 4.72 & 5.08 & 4.84 & 7.80 & 8.80 & 8.40 \\
Preoperative Education & 2.64 & 5.24 & 4.76 & 6.96 & 7.36 & 8.12 \\
Health Guidance & 6.52 & 6.96 & 6.52 & 8.32 & 9.00 & 8.96 \\
Follow-Up Plan & 4.88 & 4.28 & 5.36 & 6.20 & 7.16 & 7.68 \\
\hline
Average & 4.58 & 4.97 & 5.36 & 6.69 & 7.83 & 8.06  \\ \bottomrule
\end{tabular}
}
\caption{Detailed Human Evaluation Results for the DoctorFLAN-\textit{test}}
\label{tab:human_eval}
\end{table*}

\section{Automatic Evaluation Prompt}
\label{appendix: Evaluation Prompt}
Figure~\ref{fig:binary1} and Figure~\ref{fig:binary} show the evaluation prompt we designed for the DoctorFLAN-\textit{test} and DotaBench. 

\begin{figure*}[t]
\begin{AIbox}{Evaluation Prompt for DoctorFLAN-\textit{test}}
{\bf System Prompt:} \\
{\footnotesize
Please act as an impartial judge and evaluate the quality of the response provided by an AI assistant to the user question displayed below. \\
Requirements: Your assessment should focus primarily on the consistency between the assistant's answer and the reference answer. \\
Begin your evaluation by providing a short explanation. Be as objective as possible. After providing your explanation, you must rate the response on a scale of 1 to 10 by strictly following this format: "\lbrack \lbrack Rating \rbrack \rbrack", for example: "Rating: \lbrack \lbrack 5\rbrack \rbrack".
}
{\bf Prompt:} \\
{\footnotesize
\lbrack Question \rbrack \\
\{question\}\\
\\
\lbrack The Start of Reference Answer\rbrack \\
\{reference\}\\
\lbrack The End of Reference Answer\rbrack \\
\lbrack The Start of Assistant's Answer]\\
\{answer\}\\
\lbrack The End of Assistant's Answer \rbrack
}
\end{AIbox}
\caption{\label{fig:binary1} Evaluation Prompt for DoctorFLAN-\textit{test}.}
\end{figure*}

\label{sec:Evaluation Prompt}
\begin{figure*}[t]
\begin{AIbox}{Evaluation Prompt for DotaBench}
{\bf System Prompt:} \\
{\footnotesize
Please act as an impartial judge and evaluate the quality of the response provided by an AI assistant to the user question displayed below. \\
Requirements: Your assessment should focus on the overall quality of the responses based on the following criteria:\\
Accuracy: Evaluate the correctness and reliability of the information provided.\
Coherence: Assess the clarity and logical flow of the responses.\
Relevance: Determine how closely each response addresses the question asked.\
Thoroughness: Judge the depth and completeness of the response in covering the topic. \\
You will be given the assistant's answer and some references. The reference consists of Q\&A pairs related to the patient, which are completely accurate and can be used as a reliable source of truth.\\
You evaluation should focus on the assistant's answer to the first question. Begin your evaluation by providing a short explanation. Be as objective as possible. After providing your explanation, you must rate the response on a scale of 1 to 10 by strictly following this format: "\lbrack \lbrack Rating \rbrack \rbrack", for example: "Rating: \lbrack \lbrack 5\rbrack \rbrack".\\
}
{\bf Prompt:} \\
{\footnotesize
<|The Start of Reference|>\\
\\
\{reference\}\\
\\
<|The End of Reference|>\\
\\
\\
<|The Start of Assistant A's Conversation with User|>\\
\\
\#\#\# User:\\
\{question\_1\}\\
\\
\\
\#\#\# Assistant A:\\
\{answer\_1\}\\
\\
<|The End of Assistant A's Conversation with User|>}
\end{AIbox}

\caption{\label{fig:binary} Evaluation Prompt for DotaBench}
\end{figure*}

\section{Details of the Human Evaluation Process}
\label{appendix: Human Eval Details}
\subsection{DoctorFLAN-\textit{test}}
For the DoctorFLAN-\textit{test}, we engaged six healthcare professionals with varying levels of experience to ensure a comprehensive evaluation. The team of evaluators consisted of three mid-level professionals with 5-6 years of experience, two associate seniors each with 12 years of experience, and one senior professional with 26 years of experience. To maintain consistency in assessments, each evaluator was assigned specific questions across all six models. A total of 550 questions were distributed among the evaluators, with four handling 92 questions each, and two assessing 91 questions each, culminating in 3,300 evaluated responses. Each evaluator spent approximately six hours on this task.

\subsection{DotaBench}
For the DotaBench, which consisted of 74 data points, three doctors participated in the evaluation process. Each doctor spent three hours reviewing the data.

\subsection{Compensation}
Compensation for the evaluators was based on their professional seniority. Senior professionals were paid an hourly rate of 250 RMB, while mid-level professionals received 165 RMB per hour.

\subsection{User Interface for Human Evaluation}
\begin{figure*}[t]
\centering
\includegraphics[width=\textwidth]{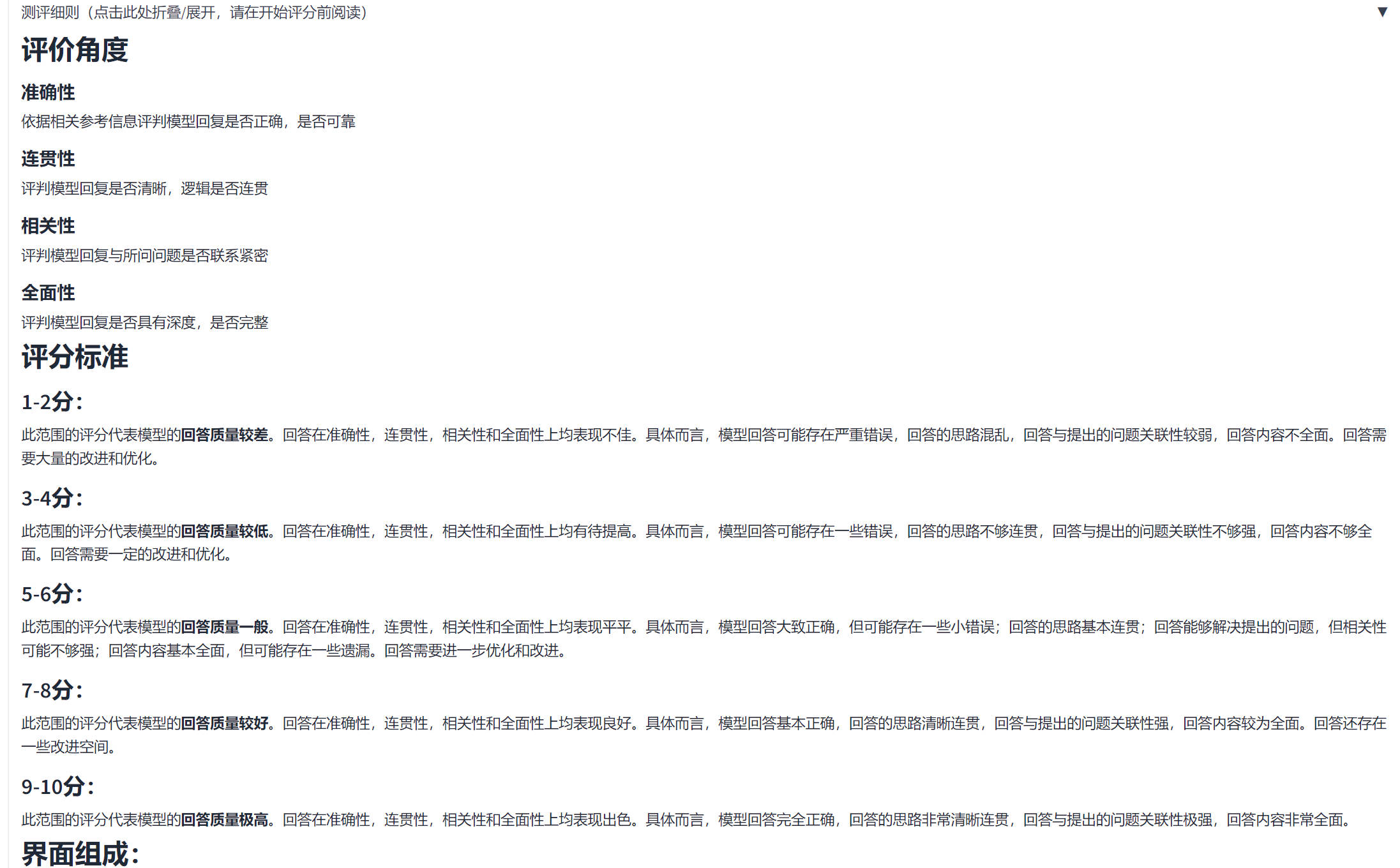}
\caption{Scoring Standards: This screenshot displays the evaluation criteria, outlining the standards and guidelines for scoring responses in the evaluation process.}
\label{fig: human_eval_web0}
\end{figure*}

\begin{figure*}[t]
\centering
\includegraphics[width=\textwidth]{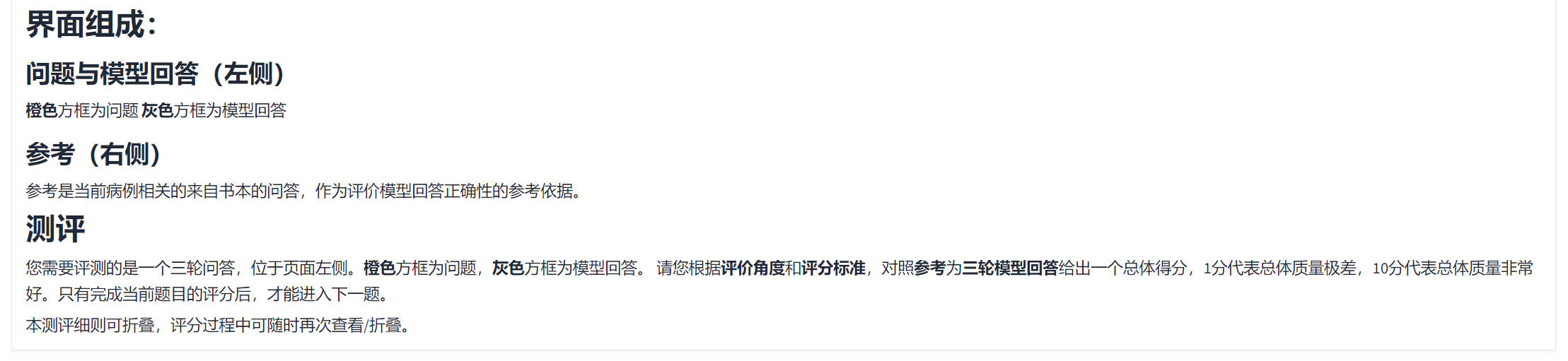}
\caption{Explanation of Website Components: This image provides a detailed explanation of the content and structure of the evaluation website, illustrating how various elements are organized for user interaction.}
\label{fig: human_eval_web1}
\end{figure*}

\begin{figure*}[t]
\centering
\includegraphics[width=\textwidth]{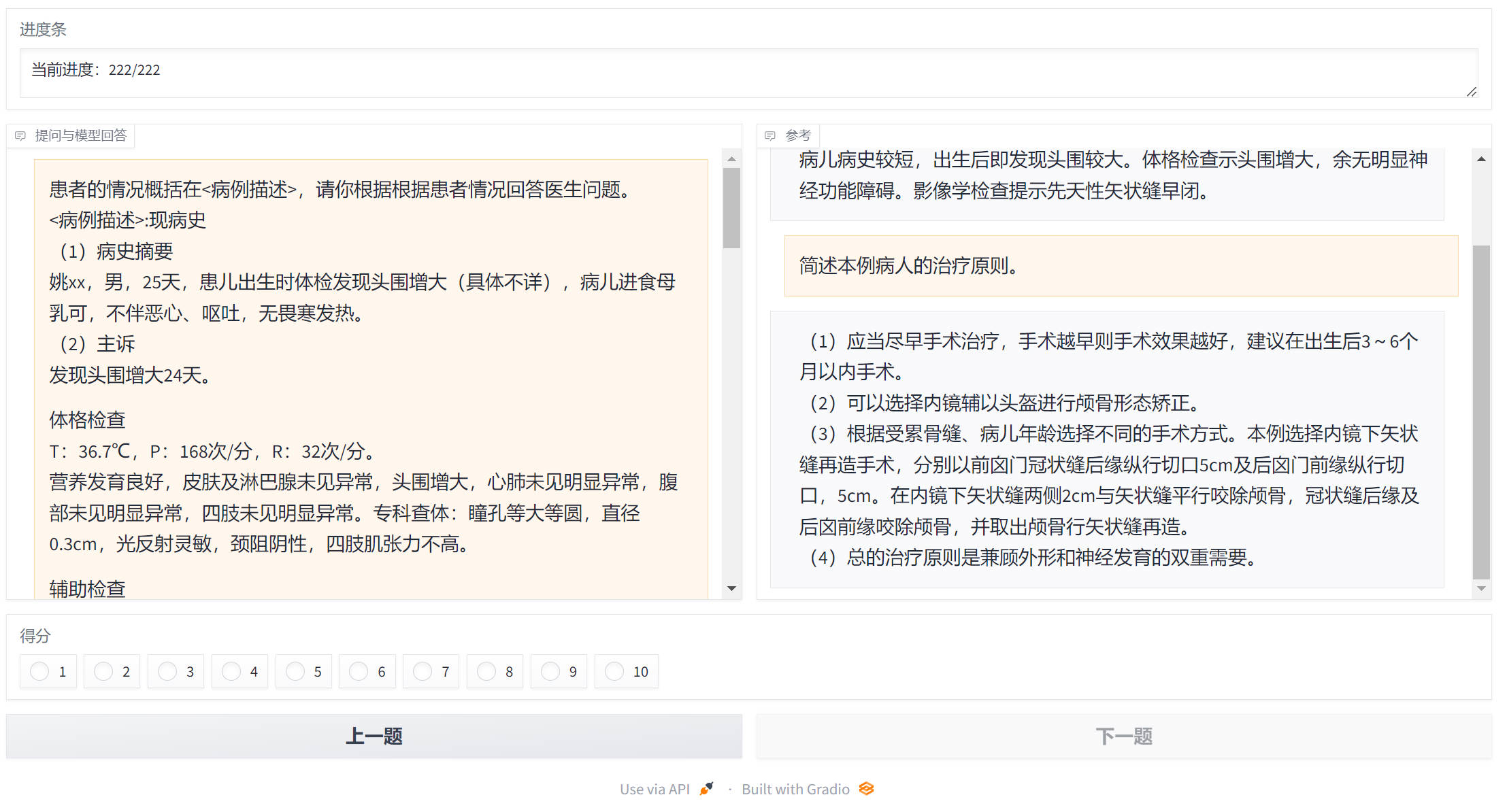}
\caption{Scoring Process: Showcases the interface for scoring, detailing how evaluators can assign scores based on the outlined criteria and submit their evaluations.}
\label{fig: human_eval_web2}
\end{figure*}

\begin{figure*}[t]
\centering
\includegraphics[width=0.8\textwidth]{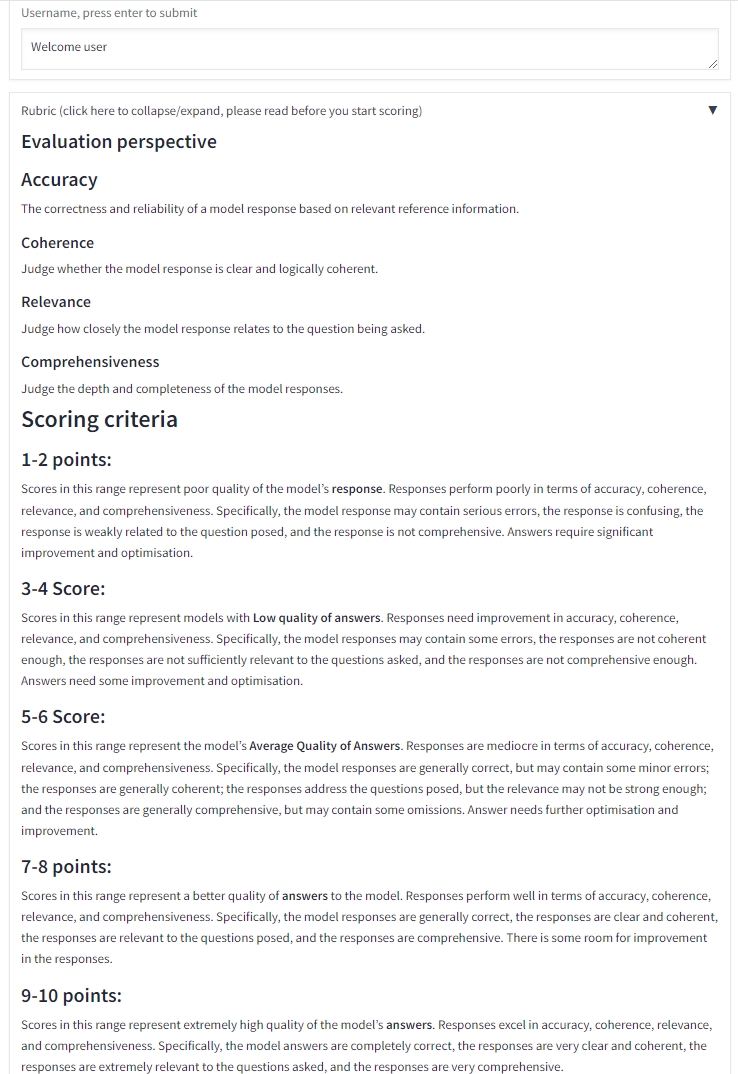}
\caption{Scoring Standards(Translated): This screenshot displays the evaluation criteria, outlining the standards and guidelines for scoring responses in the evaluation process.}
\label{fig: human_eval_web1_en}
\end{figure*}

\begin{figure*}[t]
\centering
\includegraphics[width=0.8\textwidth]{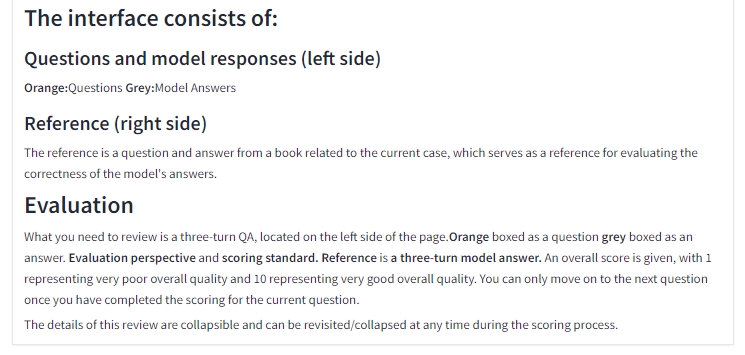}
\caption{Explanation of Website Components(Translated): This image provides a detailed explanation of the content and structure of the evaluation website, illustrating how various elements are organized for user interaction.}
\label{fig: human_eval_web2_en}
\end{figure*}

Figures \ref{fig: human_eval_web0}, \ref{fig: human_eval_web1}, \ref{fig: human_eval_web2}, \ref{fig: human_eval_web1_en} and \ref{fig: human_eval_web2_en} display screenshots of the doctor evaluation website. These figures illustrate the scoring criteria, content, and options available for evaluation.

\section{Details of Results}
\subsection{Detailed Automatic Task-by-Task Results onDoctorFLAN-\textit{test}}
\label{appendix:automatic_doctorflan_all}
Table \ref{tab:new auto eval-S} presents the scores of all models for each task.

\subsection{Detailed Automatic Turn-by-Turn Results for DotaBench}
\label{appendix:details_of_dotabench}
Table \ref{tab:DotaBench-detail-turns} presents the results for each turn across all models evaluated in the DotaBench benchmark.
\begin{table}[t]
\centering
\small
\resizebox{\linewidth}{!}{%
\begin{tabular}{lcccc}
\toprule
\multirow{3}{*}{\textbf{Model}} & \multirow{2}{*}{\textbf{Size}} & \multicolumn{3}{c}{\textbf{DotaBench}} \\
 & & \textbf{Turn 1}  & \textbf{Turn 2}  & \textbf{Turn 3} \\
\hline
\rowcolor{gray!10} \multicolumn{5}{c}{General Models} \\ \hline
Qwen-1.8B-Chat & 1.8B & 5.74\textsubscript{ $\pm$0.15}&5.81\textsubscript{ $\pm$0.42}&5.72\textsubscript{ $\pm$0.69}\\
Baichuan-13B-Chat & 13B & 8.07\textsubscript{ $\pm$0.92} & 8.54\textsubscript{ $\pm$0.13} & 7.90\textsubscript{ $\pm$0.02} \\
Baichuan2-7B-Chat&7B & 8.15\textsubscript{ $\pm$0.05} & 8.55\textsubscript{ $\pm$0.15} & 8.29\textsubscript{ $\pm$0.08} \\
Baichuan2-13B-Chat & 13B & 8.06\textsubscript{ $\pm$0.06} & 8.72\textsubscript{ $\pm$0.03} & 8.20\textsubscript{ $\pm$0.15} \\
Yi-6B-Chat&6B & 7.84\textsubscript{ $\pm$0.05}&8.36\textsubscript{ $\pm$0.04}&8.61\textsubscript{ $\pm$0.03}\\
Yi-34B-Chat&34B & 9.03\textsubscript{ $\pm$0.06}& 8.72\textsubscript{ $\pm$0.09} & 8.68\textsubscript{ $\pm$0.03}\\
\hline
\rowcolor{gray!10} \multicolumn{5}{c}{Medical Models} \\ \hline
BianQue-2 &6B& 5.60\textsubscript{ $\pm$0.07}& 4.50\textsubscript{ $\pm$0.17}&4.67\textsubscript{ $\pm$0.04} \\
DISC-MedLLM& 13B & 5.38\textsubscript{ $\pm$0.11} & 5.77\textsubscript{ $\pm$0.23} & 6.29\textsubscript{ $\pm$0.17} \\
HuatuoGPT &7B& 6.32\textsubscript{$\pm$0.24} & 6.78\textsubscript{$\pm$0.07}& 7.25\textsubscript{$\pm$0.03}\\
HuatuoGPT-II &7B& 8.43\textsubscript{ $\pm$0.06} & 8.79\textsubscript{ $\pm$0.20} & 8.44\textsubscript{ $\pm$0.13} \\
\rowcolor{gray!10} DotaGPT (Yi-6B)&6B & 9.20\textsubscript{ $\pm$0.10}&9.03\textsubscript{ $\pm$0.08}&8.91\textsubscript{ $\pm$0.10}\\
\rowcolor{gray!10} DotaGPT (Baichuan2-7B-Base) &7B& 8.96\textsubscript{ $\pm$0.07} & 9.23\textsubscript{ $\pm$0.06} & 8.82\textsubscript{ $\pm$0.16}  \\
\hline
GPT-3.5&N/A & 8.27\textsubscript{ $\pm$0.07}&8.70\textsubscript{ $\pm$0.01}&8.94\textsubscript{ $\pm$0.11}\\
Claude-3&N/A & 9.32\textsubscript{ $\pm$0.08}&9.56\textsubscript{ $\pm$0.05}&9.50\textsubscript{ $\pm$0.06}\\
GPT-4 &N/A & 9.27\textsubscript{ $\pm$0.03}&9.50\textsubscript{ $\pm$0.04}&9.47\textsubscript{ $\pm$0.04}\\ \hline
\end{tabular}
}
\caption{Detailed DotaBench Scores by Turn}
\label{tab:DotaBench-detail-turns}
\end{table}

\subsection{Detailed Human Evaluation Results for the DoctorFLAN-\textit{test}}
\label{appendix:details_of_human_doctorflan}
Table \ref{tab:human_eval} presents the human evaluation results for DoctorFLAN-\textit{test}.




\end{CJK}
\end{document}